\documentclass[lettersize,journal]{IEEEtran}
\usepackage{amsmath,amsfonts}
\usepackage{algorithmic}
\usepackage{algorithm}
\usepackage{array}

\usepackage{textcomp}
\usepackage{stfloats}
\usepackage{url}
\usepackage{verbatim}
\usepackage{graphicx}
\usepackage{cite}

\usepackage{cite}
\usepackage{amsmath,amssymb,amsfonts}
\usepackage{algorithmic}
\usepackage{graphicx}
\usepackage{textcomp}
\usepackage{caption}
\usepackage{subcaption}
\usepackage{algorithm}
\usepackage{booktabs}
\usepackage{tabularx}
\newcolumntype{Y}{>{\centering\arraybackslash}X}
\usepackage{subcaption}
\usepackage{hyperref}

\usepackage{rotating}

\usepackage{color} 

\newcommand{\INPUT}{\item[\textbf{Input:}]}
\newcommand{\OUTPUT}{\item[\textbf{Output:}]}

\begin{document}

\title{You Only Look at Once for Real-Time and Generic Multi-Task}

\author{Jiayuan Wang, \IEEEmembership{Graduate Student Member,~IEEE}, Q. M. Jonathan Wu,~\IEEEmembership{Senior Member,~IEEE},\\ and Ning Zhang,~\IEEEmembership{Senior Member,~IEEE}
\thanks{Copyright (c) 20xx IEEE. Personal use of this material is permitted. However, permission to use this material for any other purposes must be obtained from the IEEE by sending a request to pubs-permissions@ieee.org.
}
\thanks{This research was undertaken, in part, thanks to funding from the Canada Research Chairs Program, and in part by the NSERC’s CREATE program on TrustCAV. (Corresponding author: Q. M. Jonathan Wu.)}
\thanks{Jiayuan Wang, Q. M. Jonathan Wu and Ning Zhang are with the Department of Electrical and Computer Engineering, University of Windsor, Windsor, ON N9B 3P4, Canada (e-mails: wang621@uwindsor.ca, jwu@uwindsor.ca and ning.zhang@uwindsor.ca)}}

\markboth{Journal of \LaTeX\ Class Files,~Vol.~14, No.~8, August~2021}%
{Shell \MakeLowercase{\textit{et al.}}: A Sample Article Using IEEEtran.cls for IEEE Journals}

\IEEEpubid{0000--0000/00\$00.00~\copyright~2021 IEEE}

\maketitle

\begin{abstract}
High precision, lightweight, and real-time responsiveness are three essential requirements for implementing autonomous driving. In this study, we incorporate A-YOLOM, an adaptive, real-time, and lightweight multi-task model designed to concurrently address object detection, drivable area segmentation, and lane line segmentation tasks. Specifically, we develop an end-to-end multi-task model with a unified and streamlined segmentation structure. We introduce a learnable parameter that adaptively concatenates features between necks and backbone in segmentation tasks, using the same loss function for all segmentation tasks. This eliminates the need for customizations and enhances the model's generalization capabilities. We also introduce a segmentation head composed only of a series of convolutional layers, which reduces the number of parameters and inference time. We achieve competitive results on the BDD100k dataset, particularly in visualization outcomes. The performance results show a mAP50 of 81.1\% for object detection, a mIoU of 91.0\% for drivable area segmentation, and an IoU of 28.8\% for lane line segmentation. Additionally, we introduce real-world scenarios to evaluate our model's performance in a real scene, which significantly outperforms competitors. This demonstrates that our model not only exhibits competitive performance but is also more flexible and faster than existing multi-task models. The source codes and pre-trained models are released at \url{https://github.com/JiayuanWang-JW/YOLOv8-multi-task}
\end{abstract}


\section{Introduction}
\label{sec:introduction}

\begin{figure}[t!]
    \centering
    \includegraphics[width=\linewidth]{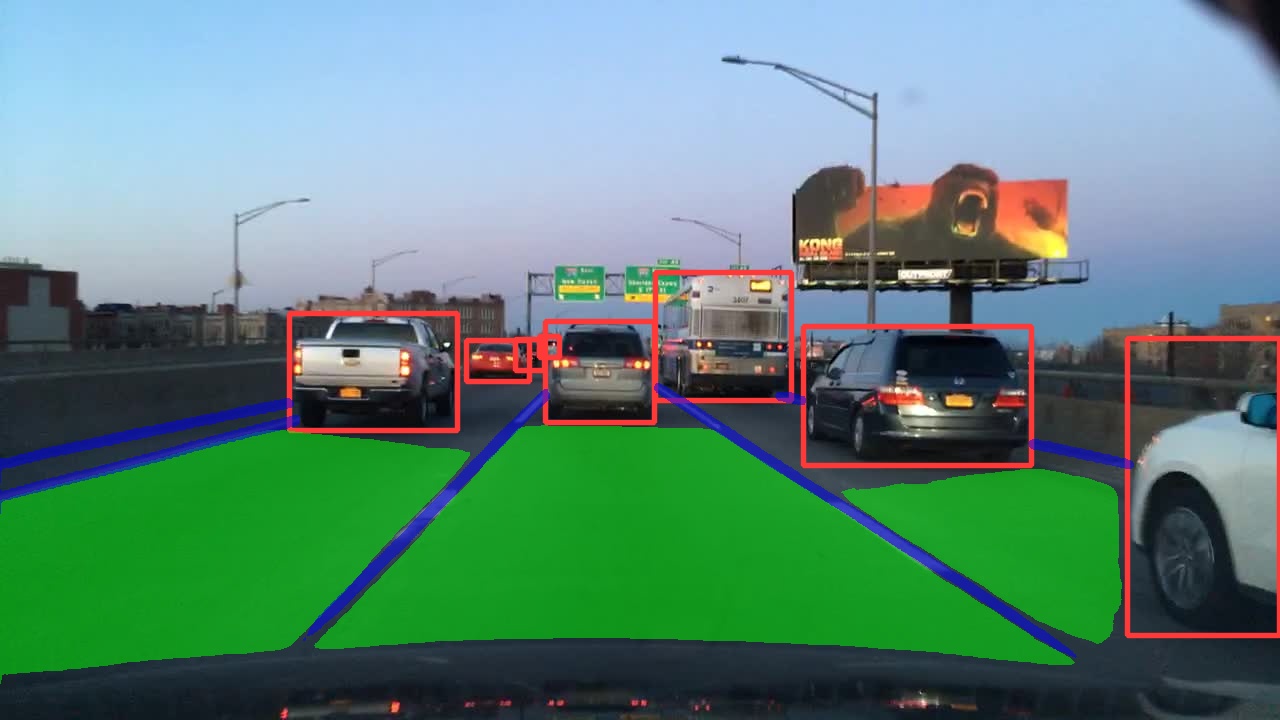}
    \caption{Multi-task in autonomous driving: object detection, drivable area segmentation and lane line segmentation}
    \label{fig:intro_fig}
\end{figure}

\IEEEPARstart{O}{wing} to rapid advancements in deep learning, the field of computer vision has flourished, particularly in autonomous driving applications~\cite{zhao2023drmnet,guo2023haze,fan2021fii}. Autonomous driving systems (ADS) provide enhanced convenience for everyday driving. Detection and segmentation are crucial for ADS, which encompasses three core tasks: object detection, drivable area segmentation, and lane line segmentation, as shown in Figure~\ref{fig:intro_fig}. In autonomous driving tasks, lidar and camera sensors are often used to acquire environmental information. However, the camera-based method stands out due to its low cost. Therefore, combining cameras with a deep learning model becomes a formidable solution. Additionally, high precision, real-time and lightweight are essential in autonomous driving tasks. In emergencies, ADS must fast and accurately decide to avoid potential collisions or safely around obstacles. Accurately and fast estimating the drivable area and lane lines is important for effective route planning. In fact, maintaining a frame rate exceeding 30 frames per second (FPS) is imperative for ADS~\cite{wang2023centernet,cai2021yolov4}. Given the limited computational capacity of edge devices, employing lightweight models is necessary. However, achieving both real-time performance and high precision simultaneously in autonomous driving through a lightweight model poses significant challenges.

\IEEEpubidadjcol

Numerous methods have been proposed for individual tasks in autonomous driving, and many of them achieve outstanding results. For object detection tasks, two primary methodologies are prevalent. The first category is two-stage techniques, epitomized by Fast R-CNN~\cite{girshick2015fast}. While these methods prioritize detection accuracy, they often come at the expense of computational efficiency. The second category is one-stage approaches, epitomized by the You Only Look Once (YOLO) series~\cite{redmon2016you,redmon2017yolo9000,redmon2018yolov3,wang2023yolov7}, ranging from version 1 to the latest, version 8. YOLOv8 is particularly noteworthy for its real-time detection capabilities, which have been widely adopted for various detection tasks. It primarily focuses on detection rather than segmentation. Although a segmentation head is introduced for segmentation tasks, the loss functions and evaluation metrics employed are mainly for the detection task. Additionally, YOLOv8 can only implement one task per model. Moreover, YOLOv8 is inherently limited in its ability to handle multiple tasks within the context of autonomous driving. Attempting to apply it under multi-task conditions necessitates the deployment of multiple models, leading to a substantial increase in both training and inference time.

For the segmentation task, one of the field's milestones is the advent of Fully Convolutional Networks (FCN)~\cite{long2015fully}. Additionally, U-Net~\cite{ronneberger2015u} and SegNet~\cite{badrinarayanan2017segnet} are common models in drivable area segmentation tasks. The drivable area typically covers a large portion of an image. Therefore, segmenting drivable areas requires information from high-level features. However, lane line segmentation differs from drivable area segmentation because of the distinct elongated and narrow features of lane lines in road images. Lane line segmentation needs low-level features for effective analysis. Recently, models such as PointLaneNet~\cite{qian2019dlt} and MFIALane~\cite{qiu2022mfialane} have gained popularity in the field of lane line segmentation. Although they achieve remarkable performance in segmentation tasks,  the integration of these tasks with detection tasks into a single model poses challenges, primarily due to the differing resolution of features required. Segmentation operates at the pixel level, whereas object detection employs grid cells in one-stage methods and utilizes selective search in two-stage approaches. While their focus differs, both segmentation and detection tasks necessitate initial feature extraction from input images. Consequently, it is possible for them to share a common backbone. Compared to using separate models for each task, integrating three distinct necks and heads into a unified model with a shared backbone can significantly save computing resources and reduce inference time.

Recently, several multi-task models have been proposed for ADS, such as YOLOP~\cite{wu2022yolop}, multi-task learning model~\cite{miraliev2023real}, Sparse U-PDP~\cite{wang2023sparse} and HybridNet~\cite{vu2022hybridnets}. All of these methods selected three tasks to construct the panoptic autonomous driving system: object detection, drivable area segmentation, and lane line segmentation, which are from the public Berkeley Deep Drive(BDD100K) dataset. Each of these methods has yielded outstanding results. Typically, their models are composed of two components: an encoder and a decoder. However, several challenges remain to be addressed. First, they have complex structures in their neck or head components, which can further impact inference times. For example, HybridNet~\cite{vu2022hybridnets} achieves outstanding performance in the detection task due to its use of an anchor-based head. This detection head performs better than the anchor-free head but tends to increase the inference time. Second, they often focus on specialized tasks when designing the loss function~\cite{wu2022yolop,miraliev2023real} or neck structure~\cite{zhao2023drmnet}. However, this approach may compromise the model's generality and necessitate considerable time for designing the structure and loss functions, as well as tuning their parameters. Therefore, developing a fast, robust, and universally applicable model is crucial.

In this work, we propose A-YOLOM, an adaptive model specifically designed for multi-task. Notably, A-YOLOM can efficiently handle multi-task using a single model with a reasonable parameter overhead. This efficiency is due to the lightweight head we design for segmentation tasks, only built by a series of convolutional layers. Compared to other state-of-the-art multi-task methods, our model has fewer total parameters. Furthermore, by employing consistent loss functions across the same task types, we maintain a uniform approach to avoid customizing for specific tasks. Importantly, we introduce an adaptive module tailored for the segmentation neck. This is primarily achieved through a learnable parameter, trained to determine whether to concatenate features of different levels. This eliminates the need for distinct structural designs for various scene tasks. To summarize, the primary contributions of our study are as follows:
\begin{itemize}
  \item We develop a lightweight model capable of integrating three tasks into a single model. This is particularly beneficial for multi-task that demand real-time processing, thereby enhancing the model's deployability on edge devices.
  \item We design a novel Adaptive Concatenate Module specifically for the neck region of segmentation architectures. This module can adaptively concatenate features without manual design and achieve a similar or better performance than well-design, further enhancing the model's generality.
  \item We design a lightweight, simple, and generic segmentation head. We adopt a unified loss function for the same type of task head, meaning we don't need to custom design for specific tasks. It is only built by a series of convolutional layers.
  \item Extensive experiments are conducted based on publicly accessible autonomous driving datasets, which demonstrate that our model can outperform existing works, particularly in terms of inference time and visualization. Moreover, we further conduct experiments using real road datasets, which also demonstrate that our model significantly outperformed the state-of-the-art approaches.
\end{itemize}

\section{Related Works}
\label{sec: Related Works}
In this section, we review existing works on detection, segmentation, and multi-task models in autonomous driving tasks. We focus our discussion on methods based on deep learning.

\subsection{Detection}
Over the past decade, the swift advancements in computer vision have significantly bolstered the progress of autonomous driving. Specifically, autonomous driving tasks can be delineated into two primary parts: detection and segmentation. The detection task encompasses object detection, identifying entities such as vehicles, pedestrians, traffic signs, etc.
The current object detection methods can be divided into two classifications~\cite{zou2023object}: two-stage and one-stage approaches. 

Two-stage methods initiate with a Region Proposal Network (RPN) to generate Regions of Interest (RoI). Subsequently, the second phase employs a deep learning network to classify these RoIs. This latter stage also fine-tunes the bounding box dimensions and position, bolstering object localization accuracy. Noteworthy examples of two-stage object detection methods include Regions with CNN features (R-CNN)~\cite{girshick2014rich}, Fast R-CNN~\cite{girshick2015fast}, and Faster R-CNN~\cite{ren2015faster}. 

In contrast, the one-stage detection approach offers an end-to-end strategy. It simultaneously predicts the bounding box and classifies the object in a single forward pass. It can be deployed on mobile devices for real-time operation and is easily implementable. The representatives of one-stage methods include You Only Look Once (YOLO)~\cite{redmon2016you,redmon2017yolo9000,redmon2018yolov3,wang2023yolov7} series, Single-Shot Multibox Detector (SSD)~\cite{liu2016ssd}, and RetinaNet~\cite{lin2017focal}. 

While one-stage methods typically lag behind two-stage methods in terms of detection performance, their real-time capabilities have made them increasingly popular in object detection tasks.

\subsection{Segmentation}
The segmentation task, including semantic and instance segmentation, differs from detection as it typically operates at the pixel level. Common segmentation tasks in autonomous driving include drivable area segmentation and lane line segmentation. However, these tasks exhibit notable differences. The drivable area typically covers a large portion of an image, but the number of regions is infrequent. As in Figure~\ref{fig:intro_fig}, only three drivable areas. Lane lines are the opposite: there are numerous in one image, yet each line is small, elongated, and narrow. 

In 2015, the deep learning model Fully Convolutional Networks (FCN)~\cite{long2015fully} was introduced for semantic segmentation tasks. With its end-to-end training capability, it achieved a 20\% improvement on the Pascal VOC 2012 dataset compared with traditional methods. However, its focus on local information leads to the loss of global details, thereby producing coarse segmentation results. To address this limitation, the encoder-decoder architecture model was proposed. Badrinarayanan et al.~\cite{badrinarayanan2017segnet} proposed SegNet, which preserves the maximum pooling indices to ensure more accurate detail restoration during the upsampling process. However, the encoder-decoder architecture has its own shortcomings. High-resolution details are often lost during the encoding process, reducing fine-grained information. This is not ideal for tasks such as lane line segmentation. Pan et al.~\cite{pan2018spatial} proposed SCNN, which employs spatial convolutions to capture the continuity and structural information in images across both horizontal and vertical directions. Due to its ability to capture spatial correlations in images, it is well-suited for detecting slender objects, such as lane lines. However, such convolution operations are computationally intensive and time-consuming. 

Given the constraints of edge devices in vehicles, it's crucial to integrate multi-tasking into a single model. This not only conserves computational resources but also meets real-time performance requirements.

\subsection{Multi-task model}
Recently, multi-task models have gained considerable attention in the research community due to their high efficiency. This is particularly important for autonomous driving, which includes various sub-tasks and often operates under the constraints of limited computational resources on edge devices. MultiNet~\cite{teichmann2018multinet} introduced an efficient, unified deep architecture capable of simultaneously addressing classification, detection, and semantic segmentation. YOLOP~\cite{wu2022yolop} extended the YOLOv5 model by introducing two additional segmentation necks and heads to perform both segmentation and detection tasks simultaneously. They separately designed the loss function for each segmentation task, which compromises the generality of the model. Sparse U-PDP~\cite{wang2023sparse} proposed a unified decoding framework that integrates three tasks into a single model. Notably, they crafted a streamlined multi-task representation through ``dynamic convolution kernels", enhanced by a dynamic interaction module which adjusts feature sampling uniquely for each task. Due to its complex decoder, it is hard to meet real-time requirements. Shokhrukh et al.~\cite{miraliev2023real} introduced a memory-efficient end-to-end framework that utilizes combinations of customized loss functions, encompassing the weighted average sum of all tasks to enhance performance. The customization and specific weighting limited the model's adaptability to new tasks. Additionally, their approach involved a sophisticated training paradigm, which could lead to overfitting and suboptimal real-world performance.

Therefore, the challenge of developing a generic and real-time multi-task model remains significant. In our work, we introduce an adaptive concatenation module and use the same loss function for tasks of the same type, further enhancing the model's generality. To decrease parameter count, we employ a lightweight backbone and a simple yet effective segmentation head, ensuring the model meets real-time requirements under limited computing resources.

\section{PROPOSED METHODOLOGY}
\label{sec: PROPOSED METHODOLOGY}
\begin{figure*}[ht]
    \centering
    \includegraphics[width=\textwidth]{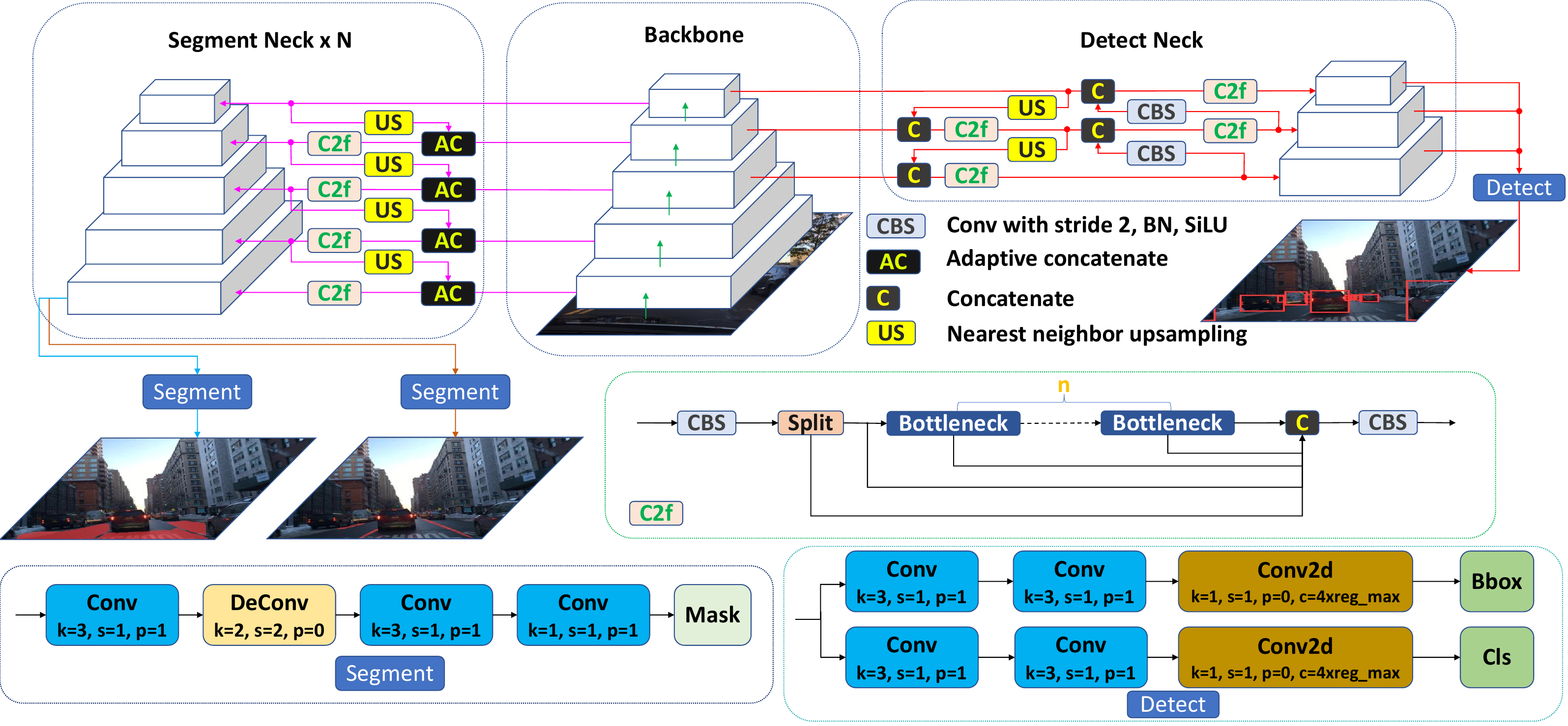}
    \caption{A-YOLOM structure}
    \label{fig:A-YOLOM_structure}
\end{figure*}

In this section, we present the details of our model, including three main components: the backbone, neck, and head. Additionally, the loss function is also included. As shown in Fig~\ref{fig:A-YOLOM_structure}. A-YOLOM model is a one-stage network with a simple encoder-decoder architecture. The encoder comprises both the backbone and neck, while the decoder consists of the head. It's worth noting that we have three necks in total: one detection neck for object detection and two segmentation necks, one for drivable areas segmentation and the other for lane lines segmentation. Our model boasts an adaptive, simple and efficient structure, which not only broadens its application range but also ensures real-time inference. 

\subsection{Encoder}
\label{subsec: Encoder}
In our work, we integrate a shared backbone network and three neck networks into a single model for the three different tasks.

\subsubsection{Backbone}
\label{subsubsec: Backbone}
The backbone comprises a series of convolutional layers to extract features from the input data. Due to the outstanding performance of YOLOv8 in detection tasks, our backbone network follows YOLOv8. Specifically, they refine CSP-Darknet53~\cite{wang2021scaled}, which serves as the backbone of YOLOv5. The key difference between this backbone and CSP-Darknet53 lies in the replacement of the c3 module with the c2f module. The c2f module combines high-level features with contextual information, further improving performance. There are also minor differences, such as using a 3x3 convolutional in place of the initial 6x6 convolutional, and the removal of the 10\textsuperscript{th} and 14\textsuperscript{th} convolutional layers. These enhancements make the backbone more effective than its predecessors in the YOLO series. 

\subsubsection{Neck}
\label{subsubsec: Neck}
The neck is responsible for fusing the features extracted from the backbone. After the backbone network, we use a Spatial Pyramid Pooling Fusion (SPPF) module to increase the receptive field and reduce computational requirements compared to SPP. These features are subsequently directed to individual necks.

In our model, we utilize three necks: one for object detection task and two for segmentation tasks, which are drivable areas and lane lines. It's worth noting that in Fig~\ref{fig:A-YOLOM_structure}, as we have two segmentation tasks, the value of N in the segment neck on the left part is equal to 2. This means we have two same segment necks for different segmentation tasks. The sky-blue and brown lines, which originate from different neck bottom layers, lead to separate segment heads. 

For the detection neck, we adopt the Path Aggregation Network (PAN)~\cite{liu2018path} structure, which includes top-down and bottom-up two Feature Pyramid Networks (FPN)~\cite{lin2017feature}. This structure merges low-level details with high-level semantic features, enriching the overall feature representation. This is invaluable for object detection, where small objects hinge on low-level features, while larger entities gain from the broader context provided by high-level features. In detection tasks, objects of varying scales are commonly present. This diversity in object size is the primary reason we selected the PAN structure for the detection neck.

For the segmentation neck, we adopt the FPN structure, renowned for its effective handling of multi-scale objects. Additionally, we introduce an adaptive concatenation module between the neck and backbone. As illustrated in Algorithm~\ref{algorithm:ACM}, input includes two feature maps: one from the previous layer neck and the other from the identical resolution backbone. Central to this module is a learnable parameter, which aims to control the concatenation of backbone features, thereby enabling adaptive modifications to the model's structure to accommodate different segmentation tasks. This module operates on each resolution level. This addition enhances the model's generality, making it easily applicable across different segmentation tasks without customization.

\begin{algorithm}[h]
\caption{Adaptive Concatenation Module}
\begin{algorithmic}[1]
\label{algorithm:ACM}
\INPUT List of tensors \( x \) includes two feature maps, x[0] from neck, x[1] from  identical resolution level backbone; channels list \( ch \)
\OUTPUT One tensor of features: \( output \)

\STATE \textbf{Initialization:}
\STATE \( dimension \leftarrow 1 \)
\STATE \( weight \leftarrow 5.0 \) \# as a learnable parameter
\STATE \( sigmoid \leftarrow \text{nn.Sigmoid()} \)
\STATE \( cv \leftarrow \text{Conv}(\text{sum}(ch), ch[0], k=1, s=1) \) 

\STATE \textbf{Forward:}
\IF{\( sigmoid(weight) > 0.5 \)}
    \STATE \( concatenated \leftarrow \text{concatenate} x \text{ along} dimension \)
    \STATE \( output \leftarrow cv(concatenated) \)
\ELSE
    \STATE \( output \leftarrow x[0] \)
\ENDIF

\RETURN \( output \)

\end{algorithmic}
\end{algorithm}

\subsection{Decoder}
\label{subsec: Decoder}
The decoder processes the feature maps from the neck to make predictions for each task. This includes predicting object classes, their corresponding bounding boxes, and masks for specific segmented objects. In our work, we employ two distinct heads: the detection head and the segmentation head. 

The detection head adopts a decoupled approach that follows the YOLOv8 detect head, using convolutional layers to convert high-dimensional features into class predictions and bounding boxes without the objectness branch. This is an anchor-free detector. The structure is shown as Detect module in Fig~\ref{fig:A-YOLOM_structure}. Note that the outputs from the detection head vary among training, validation, and prediction modes. In validation and prediction mode, three different resolutions are used as input, and output one tensor encompasses both class predictions and their respective bounding boxes. In training mode, the detection head cycle applies convolutional layers to each input, outputs three tensors in total. Each tensor contains class predictions and bounding boxes specific to its resolution. The results are subsequently used to compute the loss function. 

Our segmentation head is identical across different segmentation tasks. The structure is shown as Segment module in Fig~\ref{fig:A-YOLOM_structure}. Specifically, it comprises a series of convolutional layers that extract contextual information and one deconvolution layer to restore the resolution to the original image size. Ultimately, we obtain a pixel-level binary mask that matches the size of the original image. 0 represents the background, and 1 represents the object. Algorithm~\ref{algorithm:Segment} illustrates the processing of the segment head. Furthermore, the segment head's streamlined architecture, which comprises just 7,940 parameters, significantly bolsters the model's deployability.

\begin{algorithm}[h]
\caption{Segment Head}
\label{algorithm:Segment}
\begin{algorithmic}[1]
\INPUT Number of classes \( nc \); channels \( ch \); Feature map \( x \)
\OUTPUT Segmentation mask: \( mask \)

\STATE \textbf{Initialization:}
\STATE \( fd \leftarrow 32 \) \# intermediate convolutional feature dimension
\STATE \( cv1 \leftarrow \text{Conv}(ch[0], fd, k=3) \)
\STATE \( upsample \leftarrow \text{ConvTranspose2d}(fd, \frac{fd}{2}, 2, 2, 0) \)
\STATE \( cv2 \leftarrow \text{Conv}(\frac{fd}{2}, \frac{fd}{4}, k=3) \)
\STATE \( cv3 \leftarrow \text{Conv}(\frac{fd}{4}, nc+1) \)
\STATE \( sigmoid \leftarrow \text{nn.Sigmoid()} \)

\STATE \textbf{Forward:}
\STATE \( mask \leftarrow cv3(cv2(upsample(cv1(x)))) \)
\RETURN \( mask \)
\end{algorithmic}
\end{algorithm}

\subsection{Loss Function}
\label{subsec: Loss Function}
We employ an end-to-end training approach with a multi-task loss function. Specifically, our loss function comprises three components: one detection and two segmentation. The formula is shown as follows:
\begin{equation}
\mathcal{L}=\mathcal{L}_{\text {det}}+\mathcal{L}_{\text {segda}}+\mathcal{L}_{\text {segll}}
\end{equation}
where $\mathcal{L}_{\text {det}}$ for the object detection task, $\mathcal{L}_{\text {segda}}$ for the drivable area segmentation task, and $\mathcal{L}_{\text {segll}}$ for the lane line segmentation task. 

For detection tasks, the loss function is divided into two main branches: the classification branch and the bounding box branch. The classification branch includes binary cross-entropy loss, denoted as \( \mathcal{L}_{\text{BCE}} \). The bounding box branch includes distribution focal loss (DFL)~\cite{li2020generalized}, denoted as \( \mathcal{L}_{\text{DFL}} \) and complete IoU (CIoU) loss~\cite{zheng2020distance}, represented as \( \mathcal{L}_{\text{CIoU}} \). Thus, the detection loss $\mathcal{L}_{\text {dec}}$ can be represented as:
\begin{equation}
\mathcal{L}_{\text {det}}=\lambda_{\text {BCE}} \mathcal{L}_{\text {BCE}}+\lambda_{\text {DFL}} \mathcal{L}_{\text {DFL}}+\lambda_{\text {CIoU}} \mathcal{L}_{\text {CIoU}}
\end{equation}
where $\lambda_{\text {BCE}}$, $\lambda_{\text {DFL}}$, and $\lambda_{\text {CIoU}}$ are corresponding coefficients.
\begin{equation}
\mathcal{L}_{\text {BCE}}=-\left[y_n \log x_n+\left(1-y_n\right) \log \left(1-x_n\right)\right]
\end{equation}
where $x_n$ is the predicted classification of each object. $y_n$ is the ground truth of each object. \( \mathcal{L}_{\text{BCE}} \) measures the classification error between the predicted and the ground truth. 
\begin{equation}
\begin{gathered}
\mathcal{L}_{\mathrm{DFL}}\left(\mathcal{S}_i, \mathcal{S}_{i+1}\right)=-\left(\left(y_{i+1}-y\right) \log \left(\mathcal{S}_i\right)+\left(y-y_i\right) \log \left(\mathcal{S}_{i+1}\right)\right)\\
S_i=\frac{y_{i+1}-y}{y_{i+1}-y_i}, S_{i+1}=\frac{y_i-y}{y_i-y_{i+1}}\\
\end{gathered}
\end{equation}
where $y$ is the ground truth of the bounding box coordinate, which is a decimal. $y_{i+1}$ is ceiling of ground truth $y$. $y_{i}$ is floor of ground truth $y$. \( \mathcal{L}_{\text{DFL}} \) measures the displacement between the predicted and ground truth feature locations to make the predicted bounding boxes close to the actual ones.
\begin{equation}
\begin{gathered}
\mathcal{L}_{\text {CIoU}}=1-C I o U\\
C I o U=\operatorname{IoU}-\frac{\rho^2\left(b, b^{g t}\right)}{c^2}-\alpha v \\
v=\frac{4}{\pi^2}\left(\arctan \frac{w^{g t}}{h^{g t}}-\arctan \frac{w}{h}\right)^2 \\
\alpha=\frac{v}{(1-I o U)+v}
\end{gathered}
\end{equation}
where $b$ is the central point of the prediction box, $b^{g t}$ is the central point of the ground truth box. $\rho$ is the Euclidean distance between prediction and ground truth points. $c$ is the diagonal length of the smallest enclosing rectangle of the two boxes. Both $v$ and $\alpha$ are the coefficients used to control the ratio. $h$ is the width and height of the prediction box, $w^{g t}$ and $h^{g t}$ are the width and height of the ground truth box. \( \mathcal{L}_{\text{CIoU}} \) integrates aspects of overlap, distance, and aspect ratio consistency to measure the difference between predicted and ground truth bounding boxes. As a result, it enables the model to more precisely target the object's shape, size, and orientation.

For the segmentation tasks, we utilize an identical loss function. That means $\mathcal{L}_{\text {segda}}$ has the same formula with $\mathcal{L}_{\text {segll}}$. We will collectively refer to them as $\mathcal{L}_{\text {seg}}$. The formula is shown as follows:
\begin{equation}
\mathcal{L}_{\text {seg}}=\lambda_{\text {FL}} \mathcal{L}_{\text {FL}}+\lambda_{\text {TL}} \mathcal{L}_{\text {TL}}
\end{equation}
where $\mathcal{L}_{\text {FL}}$ and $\mathcal{L}_{\text {TL}}$ are focal loss~\cite{lin2017focal} and tversky loss~\cite{salehi2017tversky}. Both are widely used loss functions in segmentation tasks. $\lambda_{\text {FL}}$ and $\lambda_{\text {TL}}$ are corresponding coefficients. 
\begin{equation}
\mathcal{L}_{\text {FL}}=-\alpha_t\left(1-p_t\right)^\gamma \log \left(p_t\right)
\end{equation}
where $p_t$ is the probability of the model predicting the positive class. $y$ is the ground truth of each pixel. $\alpha_t$ is a weighting factor to balance the importance of positive/negative examples. $\gamma$ is a focusing parameter to modulate the contribution of each example to the loss. Focal loss offers a robust solution for handling imbalance samples, ensuring that the model doesn't become overwhelmingly biased towards the predominant and easily learned class. Instead, it places greater emphasis on challenging and underrepresented areas. 
\begin{equation}
\mathcal{L}_{\text {TL}}=1- \frac{T P}{T P+\alpha F N+\beta F P}
\end{equation}
Tversky loss is an extension of the Dice loss, introducing two additional parameters ($\alpha$ and $\beta$) to assign distinct weights to false positives and false negatives, thereby enhancing its capability to handle imbalance tasks.

\subsection{Training Paradigm}
\label{subsec: Training Paradigm}
Our training paradigm is different from the multi-task learning methods prevalent in panoptic autonomous driving tasks. We utilize an end-to-end training mode, performing backward only once for each batch. This means the entire network is optimized collectively, without freezing specific layers or alternating optimization, thus reducing the training time. Algorithm~\ref{algorithm:A-YOLOM training stage} illustrates the step-by-step training process. In each epoch, the required predicted results list $\hat{y}$ is obtained through a single forward propagation, encompassing information such as detection bounding boxes, classification information, and segmentation masks. Subsequently, each task loss is calculated and summed into a single loss $\mathcal{L}$. Then, backward propagation is performed only once to optimize the model across all tasks. After completing an epoch of training, the model is evaluated. If there's no improvement in performance compared to the last 50 epochs, training terminates prematurely. Otherwise, the training will stop after training 300 epochs. 

\begin{algorithm}[h]
\caption{A-YOLOM training stage}
\label{algorithm:A-YOLOM training stage}
\begin{algorithmic}[1]
\INPUT Target end-to-end network $F$ with parameters $\Theta$;
Training dataset $\tau$;
Validation dataset $x_{val}$;
Threshold for convergence $\phi$;
$y$ is the label list
\OUTPUT Well-trained network: $F(x; \Theta)$

\STATE Initialize the parameters $\Theta$

\FOR{epoch \( = 1 \) to \( 300 \)}
    \FOR{each batch ($x$,$y$) in $\tau$}
        \STATE \(\hat{y} \leftarrow F(x)\)  \# $\hat{y}$ is a prediction list
        \STATE \(\mathcal{L}_{\text {det}} \leftarrow (y[0],\hat{y}[0])\)
        \STATE \(\mathcal{L}_{\text {segda}} \leftarrow (y[1],\hat{y}[1])\)
        \STATE \(\mathcal{L}_{\text {segll}} \leftarrow (y[2],\hat{y}[2])\)
        \STATE \(\mathcal{L}=\mathcal{L}_{\text {det}}+\mathcal{L}_{\text {segda}}+\mathcal{L}_{\text {segll}}\)
        \STATE \(\Theta \leftarrow \arg \min _{\Theta} \mathcal{L}\)
    \ENDFOR
    \STATE \(p \leftarrow F(x_{val})\) 
    \IF {\( p = p_{\text{epoch}-50} \)}
        \RETURN \( F(x; \Theta) \) \# If the model hasn't any improvements over the last 50 epochs, then training terminates prematurely.
    \ENDIF
\ENDFOR
\RETURN \( F(x; \Theta) \)
\end{algorithmic}
\end{algorithm}

\section{EXPERIMENTS AND RESULTS}
\label{sec: EXPERIMENTS AND RESULTS}

In this section, we evaluate our model's performance and inference time on the BDD100K dataset and compare it against classical methods used in multi-task autonomous driving panoptic perception. Additionally, we present extensive ablation studies and provide an analysis of the experimental results.

\subsection{Experiment Details}
\label{subsec: Experiment Details}

\subsubsection{Dataset}
\label{subsubsec: Dataset}
The BDD100K dataset is a prominent resource in the study of autonomous driving, including 100k samples and multi-task annotations. Beyond sheer volume, the dataset's significance is underscored by its multifaceted nature, comprising diverse geographies, environmental contexts, and weather conditions. These advantages ensure that models trained on BDD100K achieve robustness and versatility, making it an ideal choice for our research. The dataset is divided into three parts: a training set with 70k images, a validation set of 10k images, and a testing set encompassing 20k images. Since the labels for the testing set are not public, we evaluate our model on the validation set. Similar to YOLOP, our detection task focuses solely on ``vehicle" detection, encompassing categories including car, bus, truck, and train. 

\subsubsection{Evaluation Metrics}
\label{subsubsec: Evaluation Metrics}
We utilize recall and mAP50 as the evaluation metrics for object detection tasks. Both metrics are widely recognized and accepted in the detection task. Recall indicates a model's capability to detect all object instances from the designated classes accurately. mAP50 is derived by taking the mean Average Precision for all classes at an IoU threshold of 0.5. Notably, Average Precision (AP) quantifies the area beneath the precision-recall curve. For the segmentation tasks, similar to YOLOP~\cite{wu2022yolop}, we utilize mIoU to evaluate the drivable area segmentation task. For the lane line segmentation task, we employ both accuracy and IoU as evaluation metrics. However, due to the number of pixel imbalances between the background and foreground in lane line segmentation, we adopt a more meaningful balance accuracy metric for evaluation. Traditional accuracy will skew results by favouring classes with a larger number of samples. In contrast, balance accuracy offers a more equitable metric by considering each class's accuracy. The formula is as follows:
\begin{equation}
\text {Line Accuracy}=\frac{\text {Sensitivity}+ \text {Specificity}}{2}
\end{equation}
where $\text{Sensitivity} = \frac{TP}{TP + FN}$ and $\text{Specificity} = \frac{TN}{TN + FP}$. 

We also compare the FPS between our model and other methods. All evaluation experiments are conducted on a GTX 1080 Ti GPU.

\subsubsection{Experimental Setup and Implementation}
\label{subsubsec: Experimental Setup and Implementation}
We compare our model with several prominent multi-task models. Due to the scarcity of multi-task models, we also include outstanding single-task or two-task models in our comparison. YOLOP and HybridNet represent the state-of-the-art and are open-source multi-task models in panoptic autonomous driving, specifically on the BDD100K dataset. Faster R-CNN and YOLOv8 are exemplary representations of two-stage and one-stage object detection networks, respectively. Both MultiNet and DLT-Net could address multiple panoptic driving perception tasks, and they have demonstrated good performance in object detection and drivable area segmentation on the BDD100k dataset. PSPNet~\cite{zhao2017pyramid} excels in semantic segmentation tasks due to its unparalleled capability to aggregate global information. Given the absence of an appropriate multi-task network handling lane line segmentation on the BDD100K dataset, we compare our model against Enet~\cite{cai2018cascade}, SCNN, and Enet-SAD~\cite{pang2019libra}, which are three leading lane detection networks. 

In order to enhance the performance, we adopt several data augmentation techniques. Specifically, we employ mosaic augmentation, which helps prevent overfitting and augment the training data. In addition, to address photometric distortions, we modify the hue, saturation, and value parameters of the images. We also incorporate fundamental augmentation techniques to handle geometric distortions such as random translation, scaling, and horizontal flipping. 

We train our model using the SGD optimizer with a learning rate ($lr$) of 0.01, momentum of 0.937, and a weight decay of 0.0005. Initially, our model undergoes a warm-up training for 3 epochs. During this warm-up phase, the momentum of the SGD optimizer is set to 0.8, and the bias lr is 0.1. In our training process, we adopt a linear learning rate annealing strategy. This strategy helps ensure that the model learns rapidly in the early stages of training and converges more stably in the later stages. Additionally, we resize the original image dimensions from 1280 × 720 to 640 × 640. For loss function coefficients, we set $\lambda_{\text {FL}}=24.0$, $\lambda_{\text {TL}}=8.0$, $\lambda_{\text {DFL}}=1.5$, $\lambda_{\text {CIoU}}=7.5$, and $\lambda_{\text {BCE}}=0.5$. We adopt $\alpha=0.7$ and $\beta=0.3$ in $\mathcal{L}_{\text {TL}}$, $\alpha_t=0.25$ and $\gamma=2$ in $\mathcal{L}_{\text {FL}}$. Finally, we train with a batch size of 120 on three RTX 4090 for 300 epochs. For evaluation, we set the confidence threshold at 0.001 and the Non-Maximum Suppression (NMS) threshold at 0.6. For prediction, the confidence threshold is set at 0.25, and the NMS threshold is at 0.45. We follow the confidence threshold and NMS settings from YOLOP. As a result, the visualization may slightly differ from the quantitative results.

\subsection{Experimental results}
\label{subsec: Experimental results}
In this subsection, we train our model end-to-end and compare its performance to other outstanding methods.

\subsubsection{Inference Time}
\label{subsubsec: Inference Time}
One of the primary challenges in deep learning applications is the inference time. Especially in autonomous driving tasks, require deploying models on edge devices, which usually have limited computational resources. Thus, ensuring models are both lightweight and real-time becomes paramount. In Table~\ref{tab:inference}, we reproduce and test the FPS of YOLOP\footnote{https://github.com/hustvl/YOLOP}, HybridNet\footnote{https://github.com/datvuthanh/HybridNets}, YOLOv8\footnote{https://github.com/ultralytics/ultralytics}, and our model, all FPS tests on a GTX 1080 Ti GPU with 1 and 32 batch sizes. The FPS calculation approach follows HybridNet. Additionally, we provide the number of parameters for each model as one of the evaluation results.

\begin{table}[h]
\centering
\caption{Comparison of different models in terms of parameters and FPS. bs is batch size.}
\label{tab:inference}
\begin{tabularx}{\linewidth}{YYYY}
\toprule
Model & Parameters & FPS (bs=1) & FPS (bs=32) \\
\midrule
YOLOP & 7.9M & 26.0 & 134.8 \\
HybridNet & 12.83M & 11.7 & 26.9 \\
YOLOv8n(det) & 3.16M & 102 & 802.9 \\
YOLOv8n(seg) & 3.26M & 82.55 & 610.49 \\
A-YOLOM(n) & 4.43M & 39.9 & 172.2 \\
A-YOLOM(s) & 13.61M & 39.7 & 96.2 \\
\bottomrule
\end{tabularx}
\end{table}

The primary difference between A-YOLOM(n) and A-YOLOM(s) is the complexity of the backbone. A-YOLOM(n) is our lightweight backbone network designed with reduced complexity, making it ideal for deployment on edge devices. A-YOLOM(s) has a more complex backbone, offering more powerful performance with increased time overhead, especially in scenarios with bs=32. Compared to other SOTA multi-task models and single-task models in panoptic autonomous driving, A-YOLOM(n) stands out as more lightweight and offers more efficiency. Specifically, compared to YOLOP, A-YOLOM(n) has significantly fewer parameters and a higher FPS. It achieves a speed-up of x1.53 at bs=1 and x1.28 at bs=32. That indicates our model is more efficient. Compared to HybridNet, A-YOLOM(n) has a significant advantage in parameters and speed. Despite having more parameters than HybridNet, A-YOLOM(s) is faster. HybridNet doesn't meet the requirements for real-time performance in both bs=1 and bs=32. We believe the longer inference time for HybridNet is due to its being an anchor-based method. Generating numerous anchor boxes will increase computational overhead, consequently reducing the overall speed of inference. In this case, we won't compare its performance with ours in the subsequent section. YOLOv8 is a single-task model, meaning it can only implement one task within a single model. This is the reason we have listed two versions: YOLOv8n(det) and YOLOv8n(seg). Although both of them are faster than other models, including ours, they require deploying three separate models on edge devices, which means combining one detection model with two segmentation models into one edge device, amounting to 9.68M parameters. This is x2.18 more than A-YOLOM(n), putting significant pressure on edge devices. Additionally, YOLOv8's performance across the three tasks is much inferior to ours, which we will discuss in the upcoming section.

\subsubsection{Multi-task Result}
\label{subsubsec: Multi-task Result}
This part presents the results of multi-task experiment, encompassing object detection, drivable area segmentation, and lane line segmentation.

We follow the YOLOP settings in the detection task, combining the car, bus, truck, and train into a ``vehicle" classification. The comparison results are shown in Table~\ref{tab:object_detect}. Based on the quantitative results, both our models achieve the best performance in terms of mAP50. This indicates that our model's accuracy is excellent in predicting detected targets. Especially when compared to MultiNet and Faster R-CNN. Additionally, A-YOLOM(n) has a complexity level for its backbone, detection neck, and head that is comparable to YOLOv8n(det). However, A-YOLOM(n) exhibits significant improvement over YOLOv8n(det) in both recall and mAP50. This demonstrates that in multi-task learning, various tasks can implicitly assist and further enhance the performance of individual tasks. On another note, YOLOv8n(det) outperforms YOLOv5s because the 's' scale model usually has a more complex backbone to enhance performance. One issue with our model is the unsatisfactory recall. YOLOP and DLT-Net achieve better recall than our model. We believe this is due to the high weight of bounding box loss in $\mathcal{L}_{det}$. This means our model is more conservative and focuses on achieving a higher mAP at the expense of recall performance. We believe that mAP50 better reflects the comprehensive performance in the detection task. Besides, our model achieves real-time performance, while YOLOP cannot when the batch size is set to 1 according to Table~\ref{tab:inference}. 

\begin{table}[h]
    \centering
    \caption{Traffic object detection results}
    \label{tab:object_detect}
    \begin{tabularx}{\linewidth}{YYY}
        \toprule
        Model & Recall (\%) & mAP50 (\%) \\
        \midrule
        MultiNet & 81.3 & 60.2 \\
        DLT-Net & \textbf{89.4} & 68.4 \\
        Faster R-CNN & 81.2 & 64.9 \\
        YOLOv5s & 86.8 & 77.2 \\
        YOLOv8n(det) & 82.2 & 75.1 \\
        YOLOP & 88.6 & 76.5 \\
        A-YOLOM(n) & 85.3 & 78.0\\
        A-YOLOM(s) & 86.9 & \textbf{81.1} \\
        \bottomrule
    \end{tabularx}
\end{table}

For the drivable area segmentation task, Table~\ref{tab:da_results} provides the quantitative results. Our model achieves the second and third-best performances in terms of mIoU. YOLOP outperforms our model because they customized the loss function for the segmentation task. In terms of mIoU, A-YOLOM(n) and A-YOLOM(s) trail YOLOP by 1\% and 0.5\% respectively. We believe this result is acceptable. While our model might exhibit a slight sacrifice in performance, it is more flexible and faster. Additionally, our model performs much better than other models, such as YOLOv8n(seg), MultiNet, and PSPNet. It's worth noting that while YOLOv8n(seg) is faster than our model, its deployment cost is higher, and its performance is significantly inferior to ours.

\begin{table}[h]
    \centering
    \caption{Drivable area segmentation results}
    \label{tab:da_results}
    \begin{tabularx}{\linewidth}{YY}
        \toprule
        Model & mIoU (\%) \\
        \midrule
        MultiNet & 71.6 \\
        DLT-Net & 72.1 \\
        PSPNet & 89.6 \\
        YOLOv8n(seg) & 78.1 \\
        YOLOP & \textbf{91.6} \\
        A-YOLOM(n) & 90.5\\
        A-YOLOM(s) & 91.0 \\
        \bottomrule
    \end{tabularx}
\end{table}

For the lane line segmentation task, Table~\ref{tab:ll_results} provides the quantitative results. A-YOLOM(s) achieves the best performance in terms of both accuracy and IoU. Specifically, A-YOLOM(s) delivers competitive results in accuracy and higher 2.3\% in IoU compared to YOLOP. It's worth noting that our model maintains the same structure and loss function across all segmentation tasks, eliminating the need for adjustments when encountering a new segmentation task. Additionally, YOLOv8 (seg) is slightly inferior to ours. Based on results in Table~\ref{tab:da_results} and Table~\ref{tab:ll_results}, we believe that our proposed neck and head are better suited for the segmentation task compared to YOLOv8 (seg). Since we use balance accuracy for evaluation, Enet, SCNN, and ENet-SAD employ different accuracy calculation approaches. Therefore, we cannot directly compare our accuracy with theirs. However, our results in IoU metrics are significantly better than theirs. This alone sufficiently demonstrates the superior performance of our model compared to theirs.

\begin{table}[h]
    \centering
    \caption{Lane line segmentation results}
    \label{tab:ll_results}
    \begin{tabularx}{\linewidth}{YYY}
        \toprule
        Model & Accuracy (\%) & IoU (\%) \\
        \midrule
        Enet & N/A & 14.64 \\
        SCNN & N/A & 15.84 \\
        ENet-SAD & N/A & 16.02 \\
        YOLOv8n(seg) & 80.5 & 22.9 \\
        YOLOP & 84.8 & 26.5 \\
        A-YOLOM(n) & 81.3 & 28.2 \\
        A-YOLOM(s) & \textbf{84.9} & \textbf{28.8} \\
        \bottomrule
    \end{tabularx}
\end{table}

\subsubsection{Visualization}
\label{subsubsec: Visualization}
\begin{figure*}[!h]
    \centering

    \begin{subfigure}[b]{0.05\textwidth}
        \centering
        \rotatebox{90}{YOLOP}
        \vspace{0.6cm} %
    \end{subfigure}%
    \begin{subfigure}{0.25\textwidth}
        \centering
        \includegraphics[width=\linewidth]{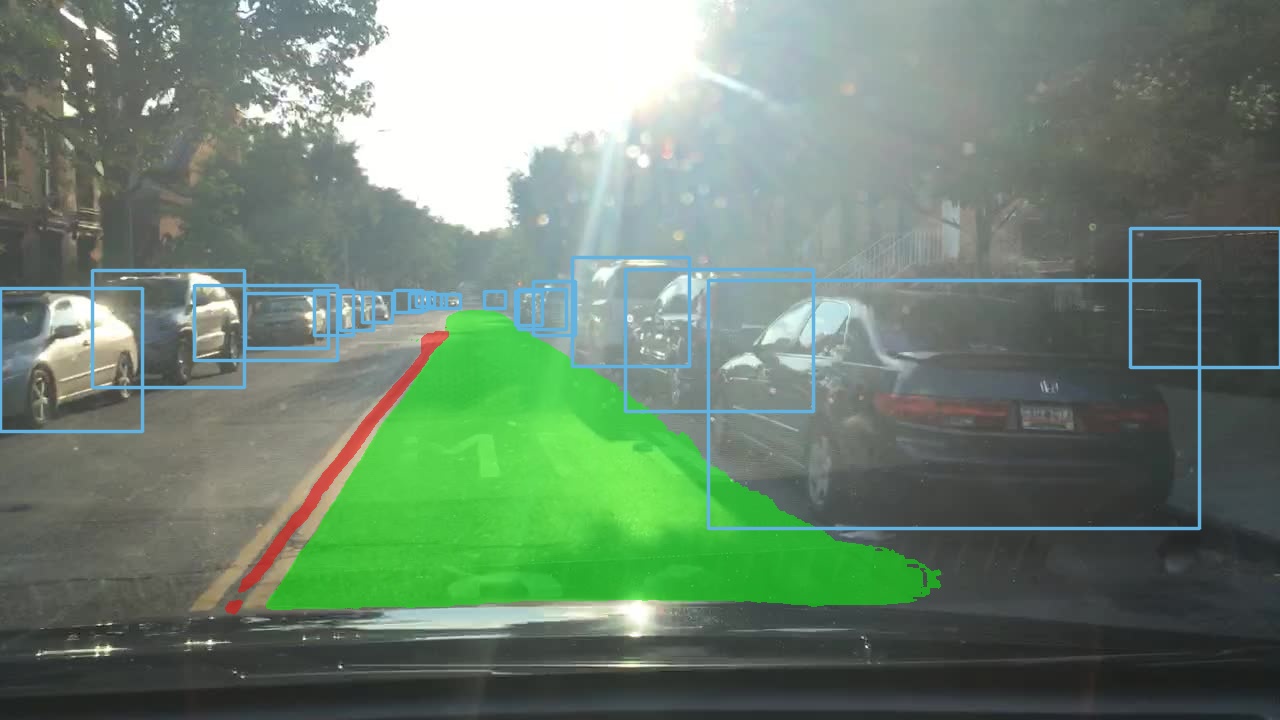}
    \end{subfigure}%
    \hspace{0.5cm}
    \begin{subfigure}{0.25\textwidth}
        \centering
        \includegraphics[width=\linewidth]{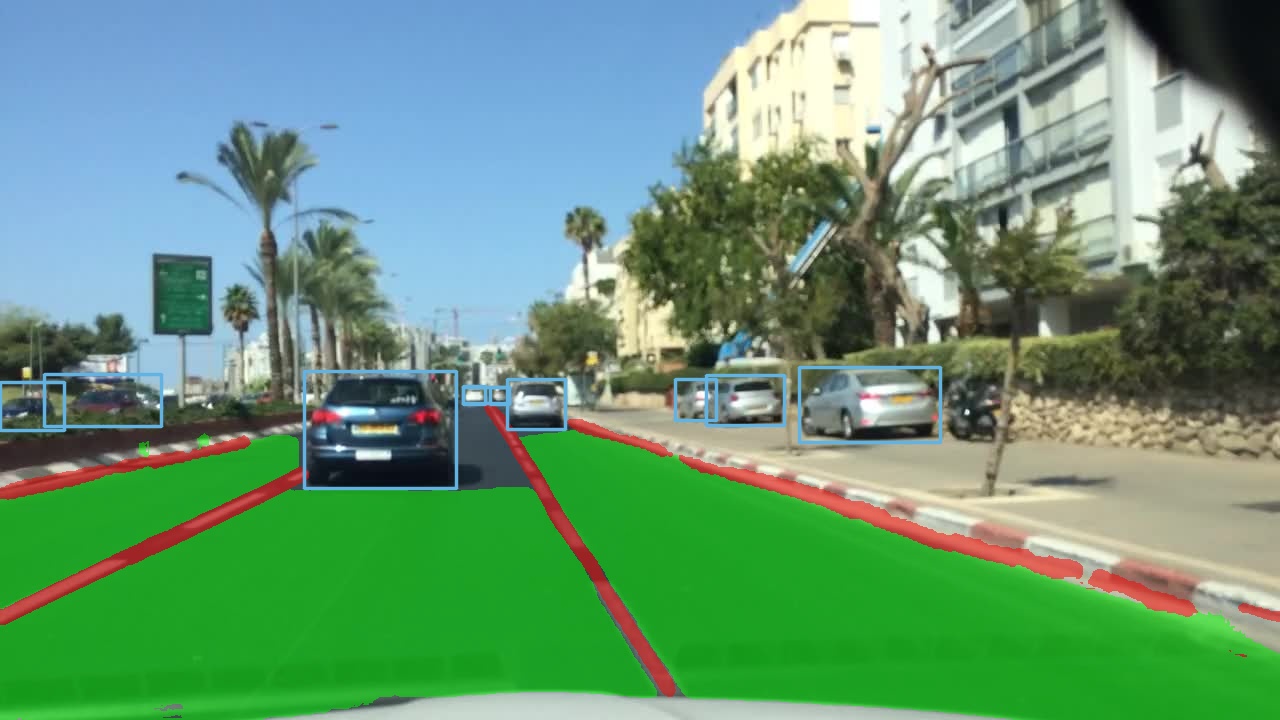}
    \end{subfigure}%
    \hspace{0.5cm}
    \begin{subfigure}{0.25\textwidth}
        \centering
        \includegraphics[width=\linewidth]{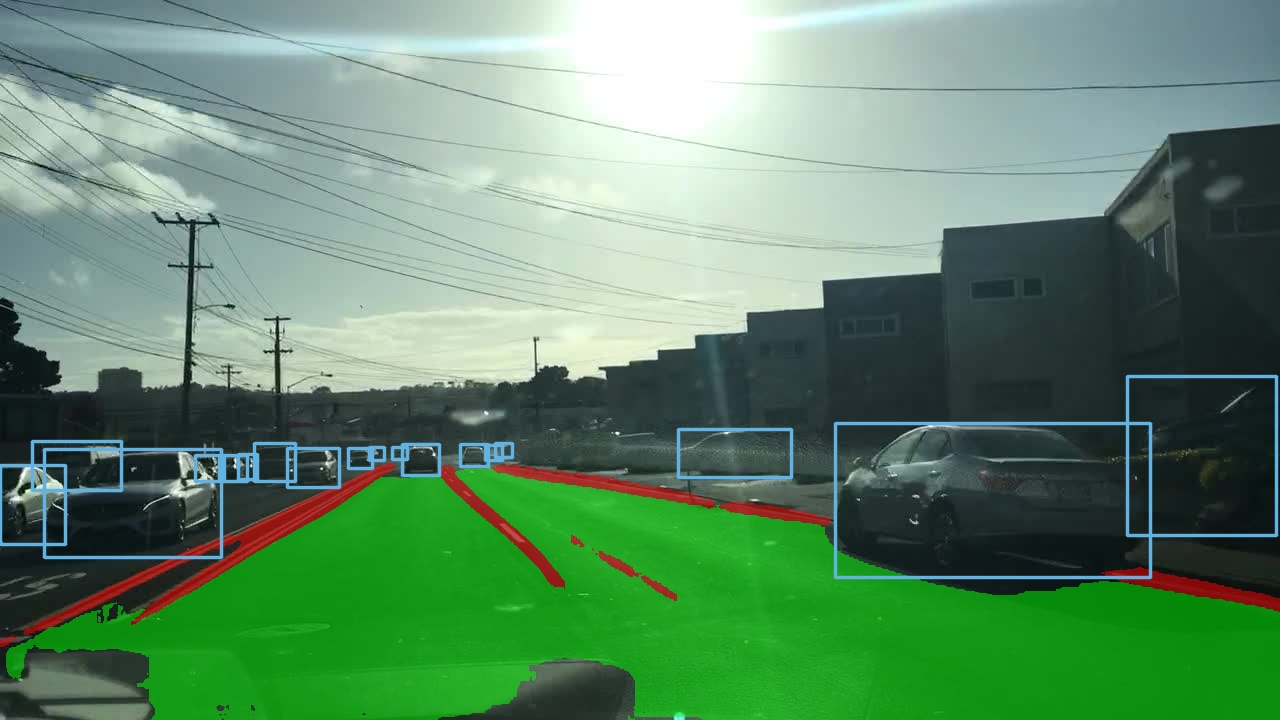}
    \end{subfigure}

    \medskip

    \begin{subfigure}[b]{0.05\textwidth}
        \centering
        \rotatebox{90}{A-YOLOM(n)}
        \vspace{0.2cm}
    \end{subfigure}%
    \begin{subfigure}{0.25\textwidth}
        \centering
        \includegraphics[width=\linewidth]{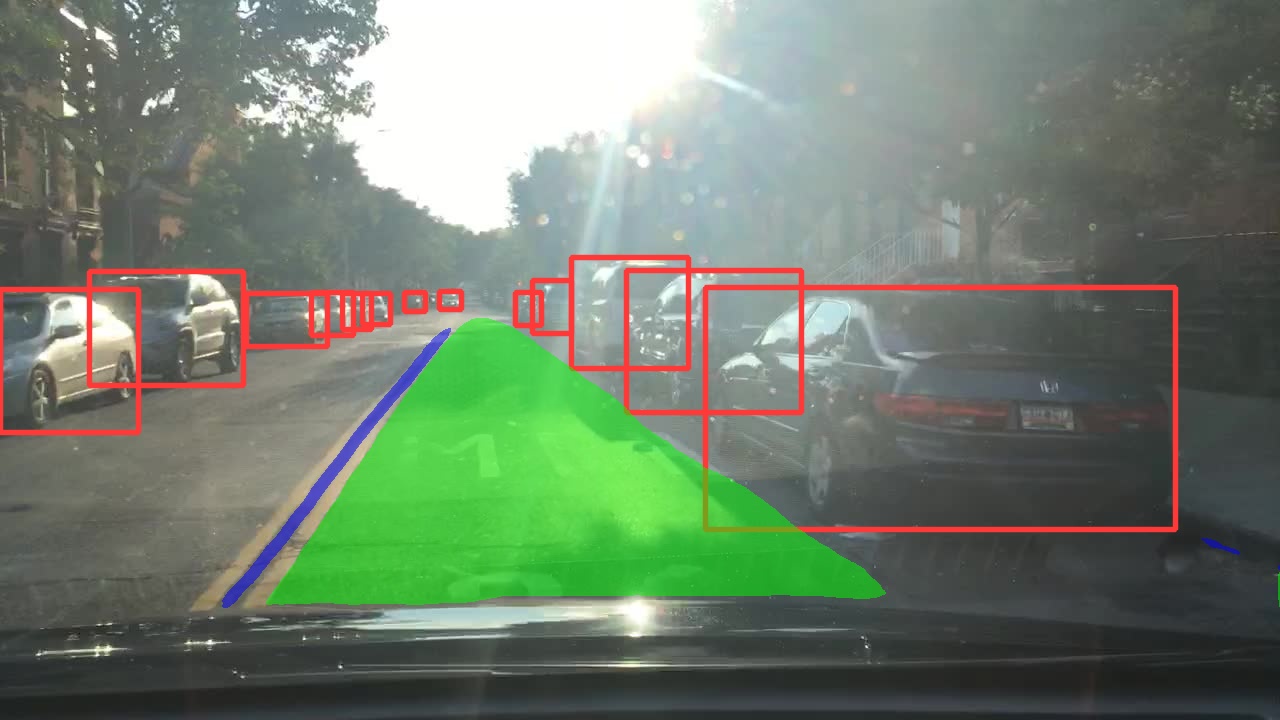}
    \end{subfigure}%
    \hspace{0.5cm}
    \begin{subfigure}{0.25\textwidth}
        \centering
        \includegraphics[width=\linewidth]{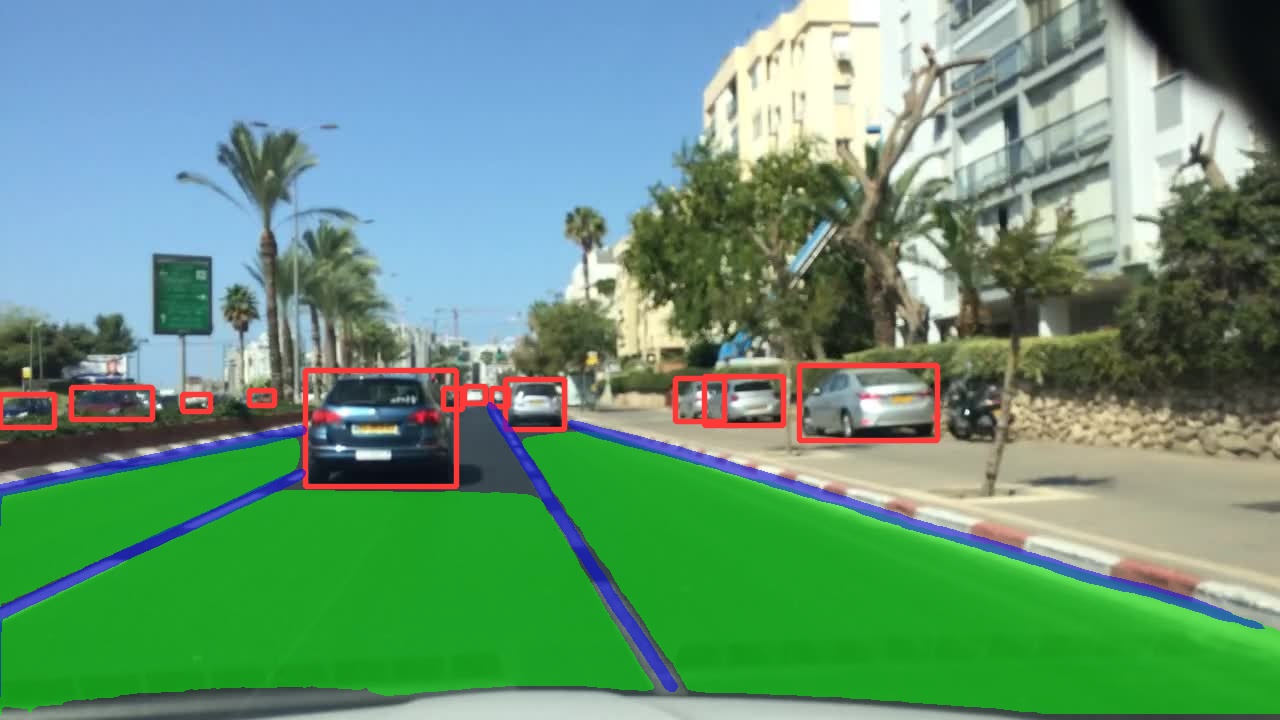}
    \end{subfigure}%
    \hspace{0.5cm}
    \begin{subfigure}{0.25\textwidth}
        \centering
        \includegraphics[width=\linewidth]{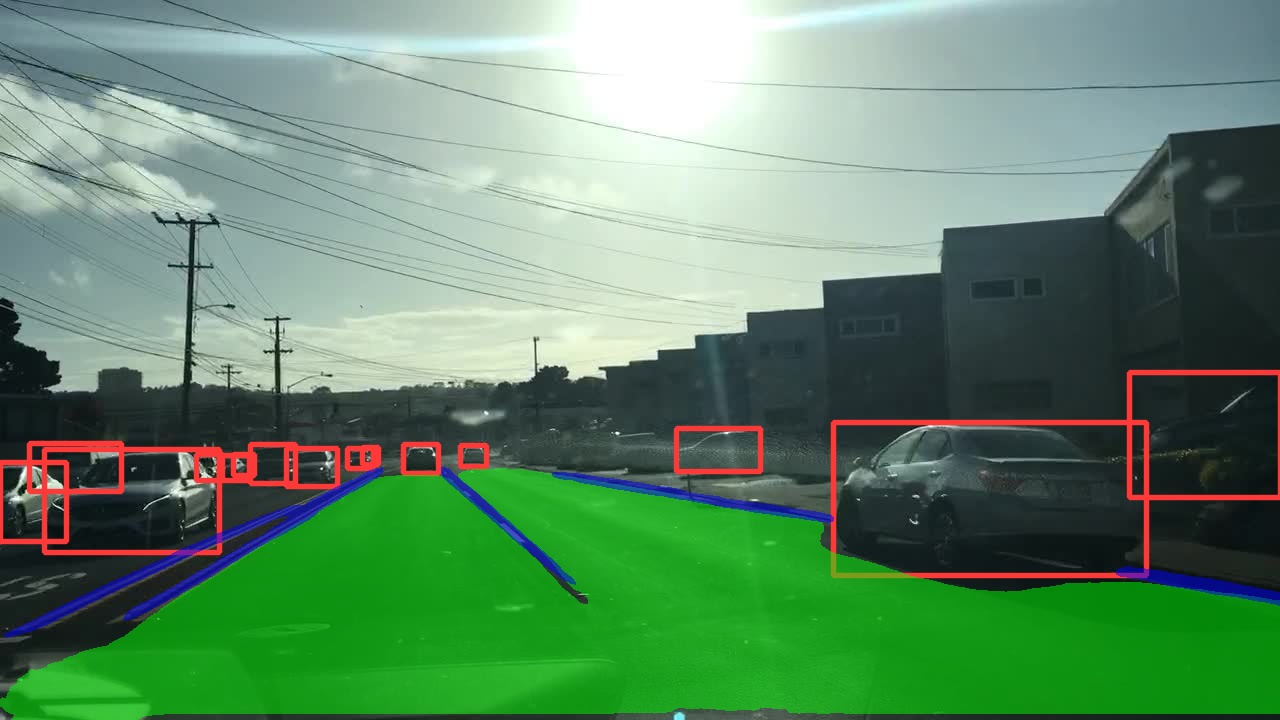}
    \end{subfigure}

    \medskip

    \begin{subfigure}[b]{0.05\textwidth}
        \centering
        \rotatebox{90}{ A-YOLOM(s)}
        \vspace{0.1cm}
    \end{subfigure}%
    \begin{subfigure}{0.25\textwidth}
        \centering
        \includegraphics[width=\linewidth]{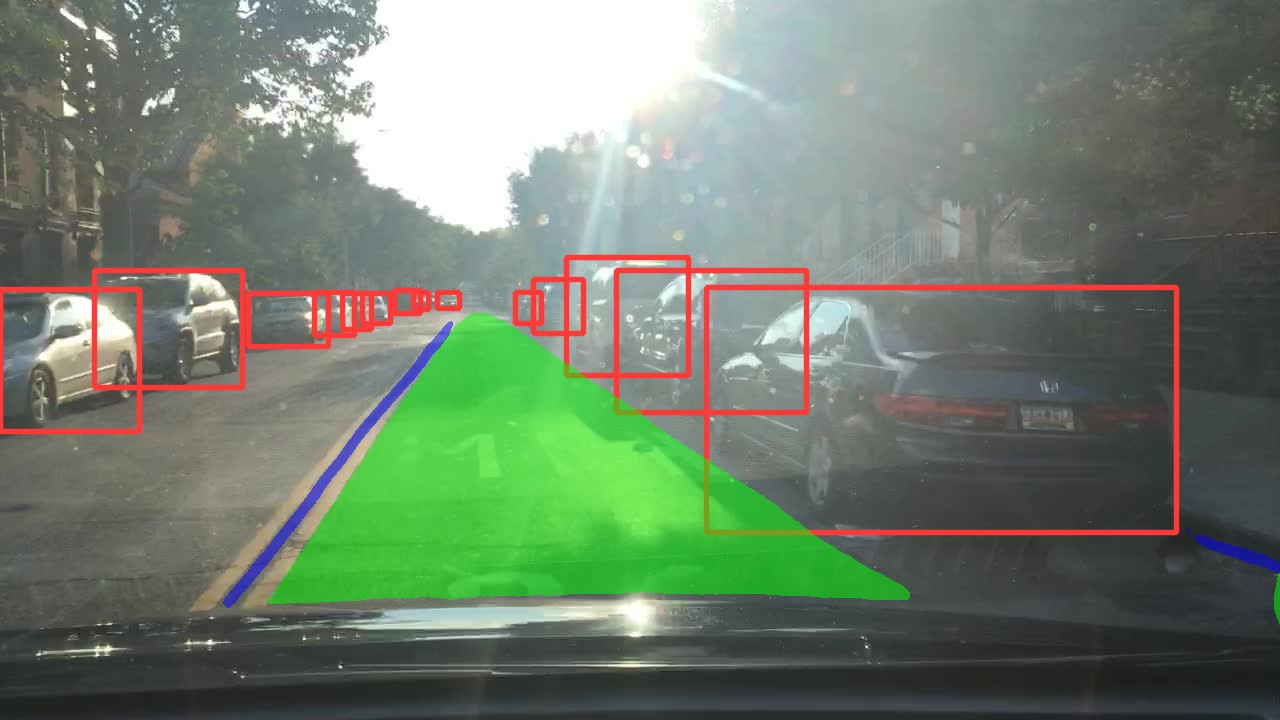}
    \end{subfigure}%
    \hspace{0.5cm}
    \begin{subfigure}{0.25\textwidth}
        \centering
        \includegraphics[width=\linewidth]{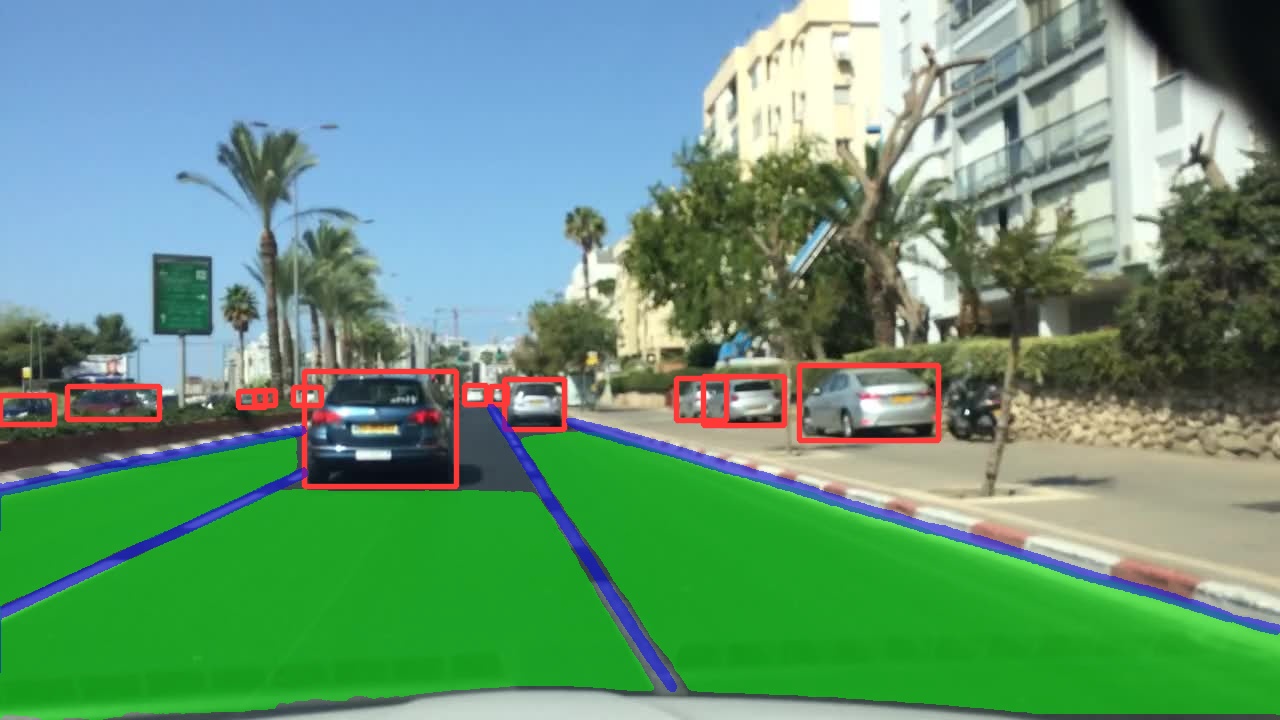}
    \end{subfigure}%
    \hspace{0.5cm}
    \begin{subfigure}{0.25\textwidth}
        \centering
        \includegraphics[width=\linewidth]{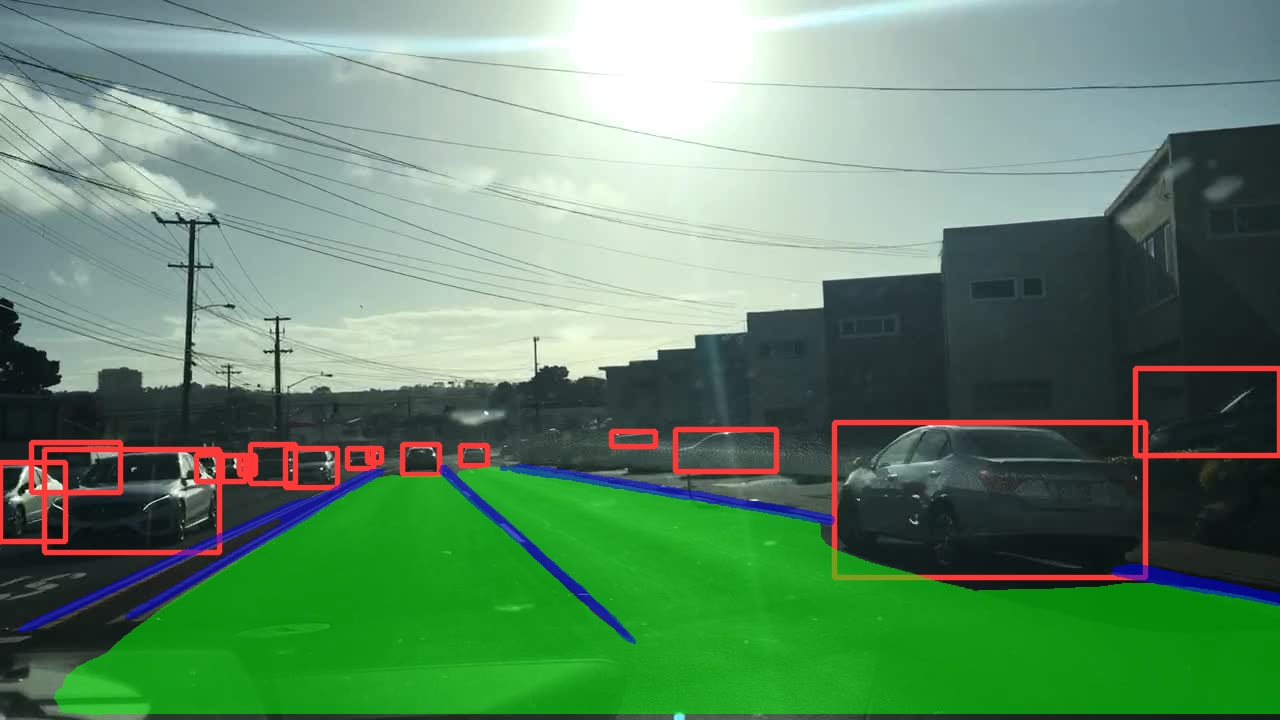}
    \end{subfigure}

    \caption{Visual Comparison of Results on a Sunny Day}
    \label{fig:sunny_day_comparison}
\end{figure*}
This section presents a visual comparison between YOLOP and our model. We evaluate performance not only in favourable weather conditions but also under adverse conditions, including strong sunlight, nighttime, rain, and snow. We will analyze each scene one by one.

Figure~\ref{fig:sunny_day_comparison} displays the results from a sunny day. As we know, strong sunlight can affect a driver's vision. Similarly, it impacts image acquisition from the camera, further affecting the model's performance. This is a challenge for the drivable area and lane line segmentation. In this challenging scenario, our model outshines YOLOP. Specifically, under the conditions of intense sunlight (as seen in the left and right images), our model delivers accurate lane line predictions and provides smoother indications of drivable areas. Additionally, our model exhibits greater accuracy in detecting smaller and farther vehicles. As observed in the middle and right images, our model successfully detects vehicles located far in the distance, both on the road in opposite driving directions and near the house. We believe that our model outperformed YOLOP in this scene according to the visualization results.

\begin{figure*}[!h]
    \centering

    \begin{subfigure}[b]{0.05\textwidth}
        \centering
        \rotatebox{90}{YOLOP}
        \vspace{0.6cm} 
    \end{subfigure}%
    \begin{subfigure}{0.25\textwidth}
        \centering
        \includegraphics[width=\linewidth]{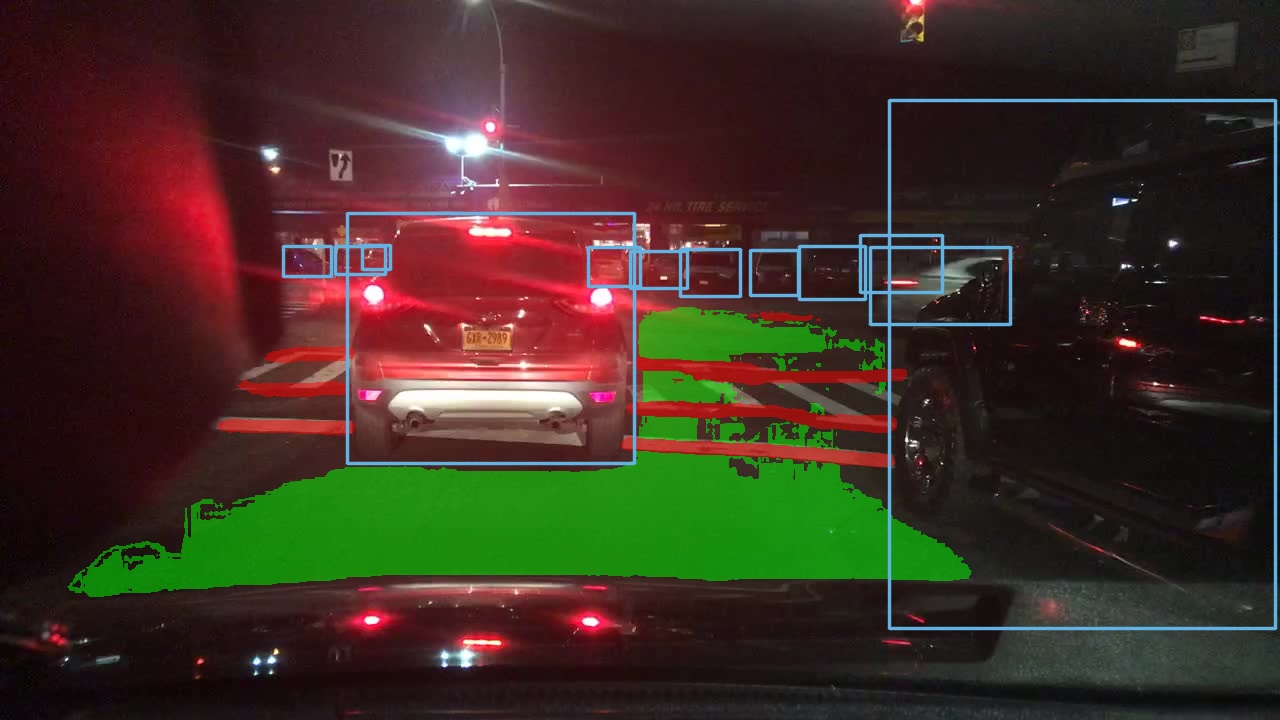}
    \end{subfigure}%
    \hspace{0.5cm}
    \begin{subfigure}{0.25\textwidth}
        \centering
        \includegraphics[width=\linewidth]{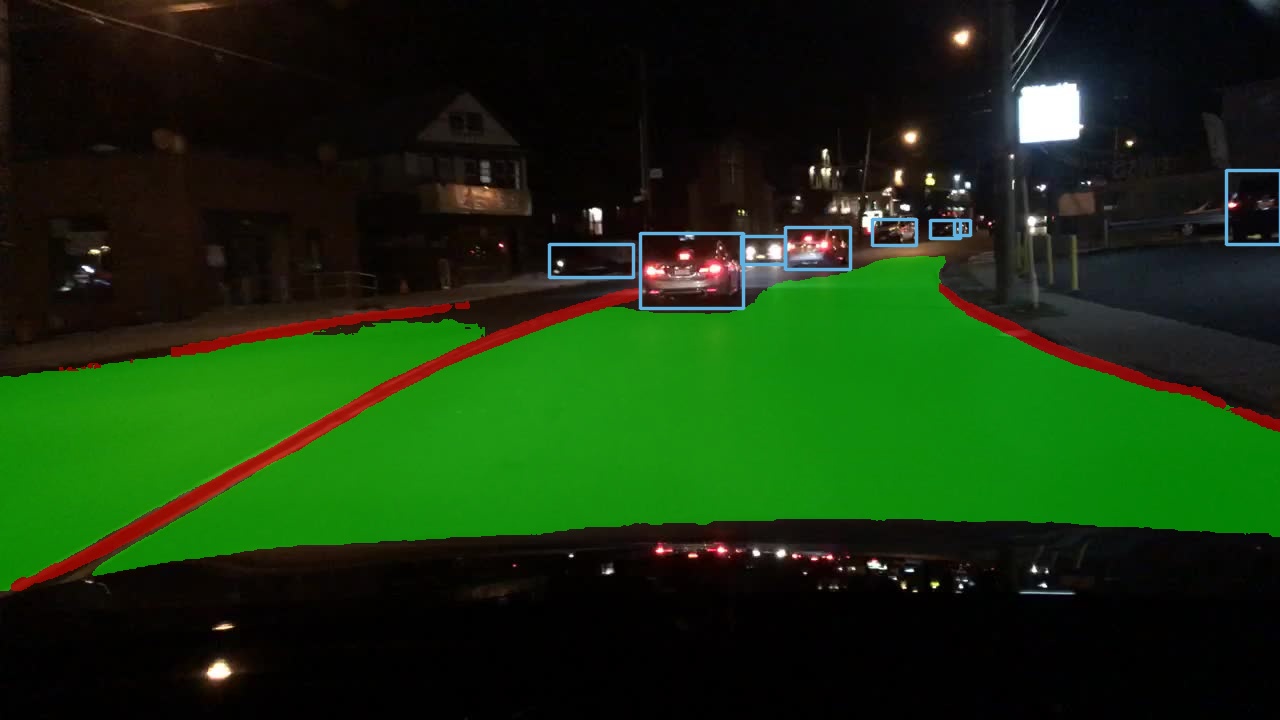}
    \end{subfigure}%
    \hspace{0.5cm}
    \begin{subfigure}{0.25\textwidth}
        \centering
        \includegraphics[width=\linewidth]{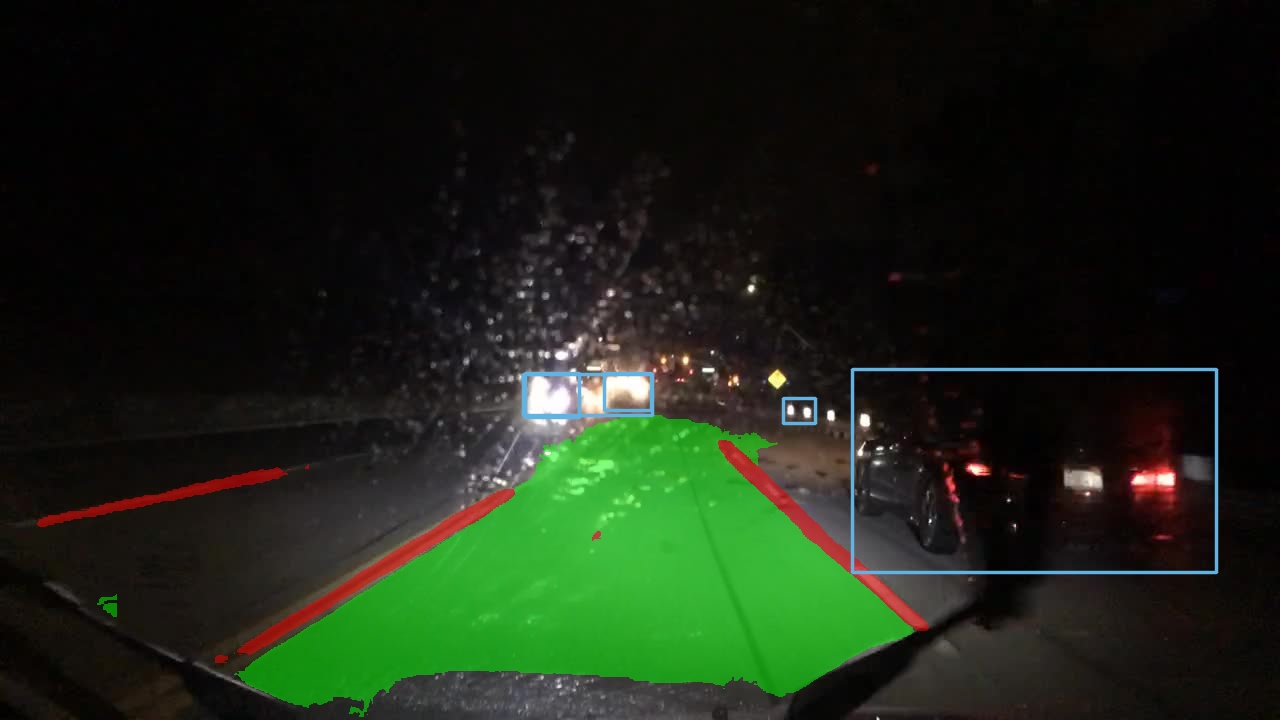}
    \end{subfigure}
    
    \medskip
    
    \begin{subfigure}[b]{0.05\textwidth}
        \centering
        \rotatebox{90}{A-YOLOM(n)}
        \vspace{0.2cm}
    \end{subfigure}%
    \begin{subfigure}{0.25\textwidth}
        \centering
        \includegraphics[width=\linewidth]{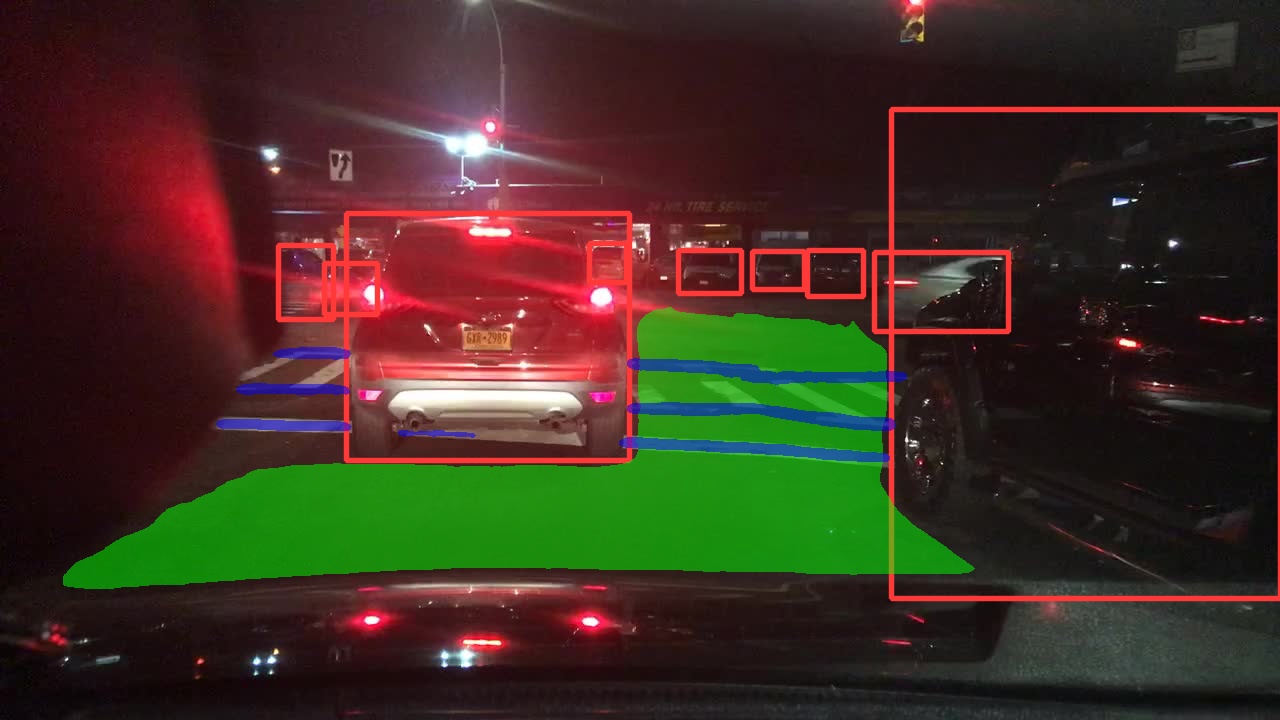}
    \end{subfigure}%
    \hspace{0.5cm}
    \begin{subfigure}{0.25\textwidth}
        \centering
        \includegraphics[width=\linewidth]{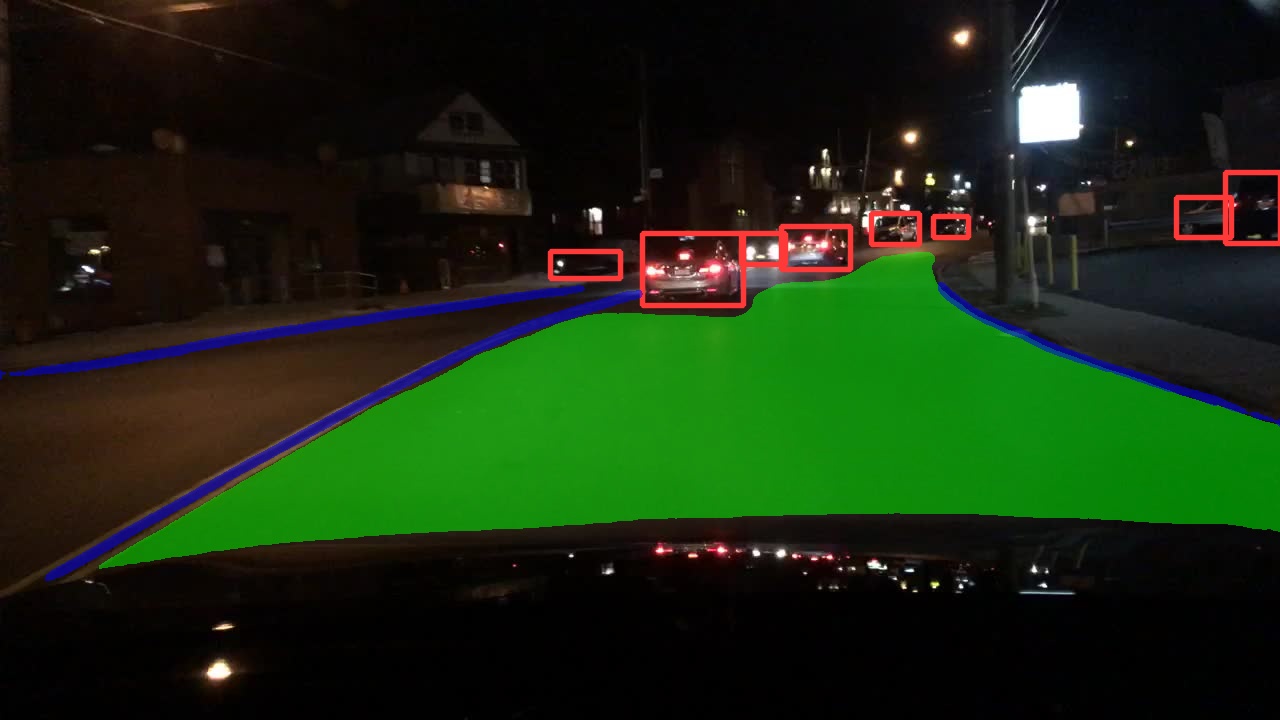}
    \end{subfigure}%
    \hspace{0.5cm}
    \begin{subfigure}{0.25\textwidth}
        \centering
        \includegraphics[width=\linewidth]{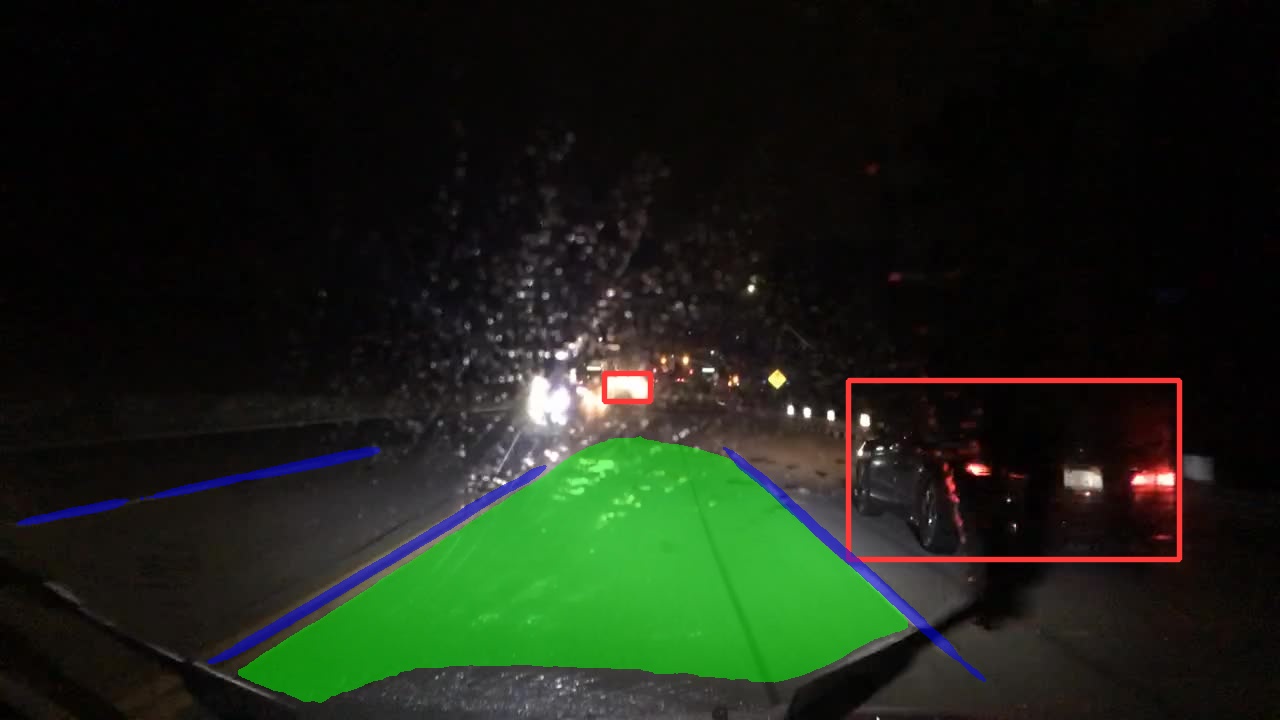}
    \end{subfigure}

    \medskip
    
    \begin{subfigure}[b]{0.05\textwidth}
        \centering
        \rotatebox{90}{ A-YOLOM(s)}
        \vspace{0.1cm}
    \end{subfigure}%
    \begin{subfigure}{0.25\textwidth}
        \centering
        \includegraphics[width=\linewidth]{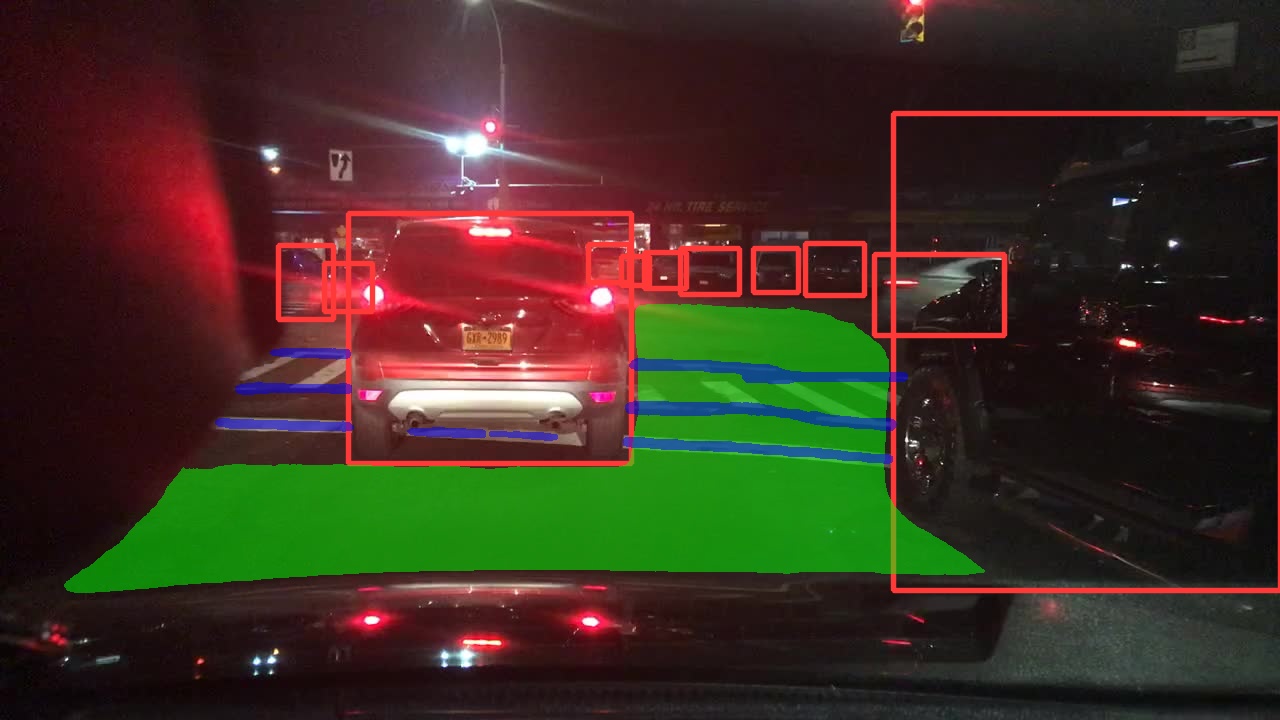}
    \end{subfigure}%
    \hspace{0.5cm}
    \begin{subfigure}{0.25\textwidth}
        \centering
        \includegraphics[width=\linewidth]{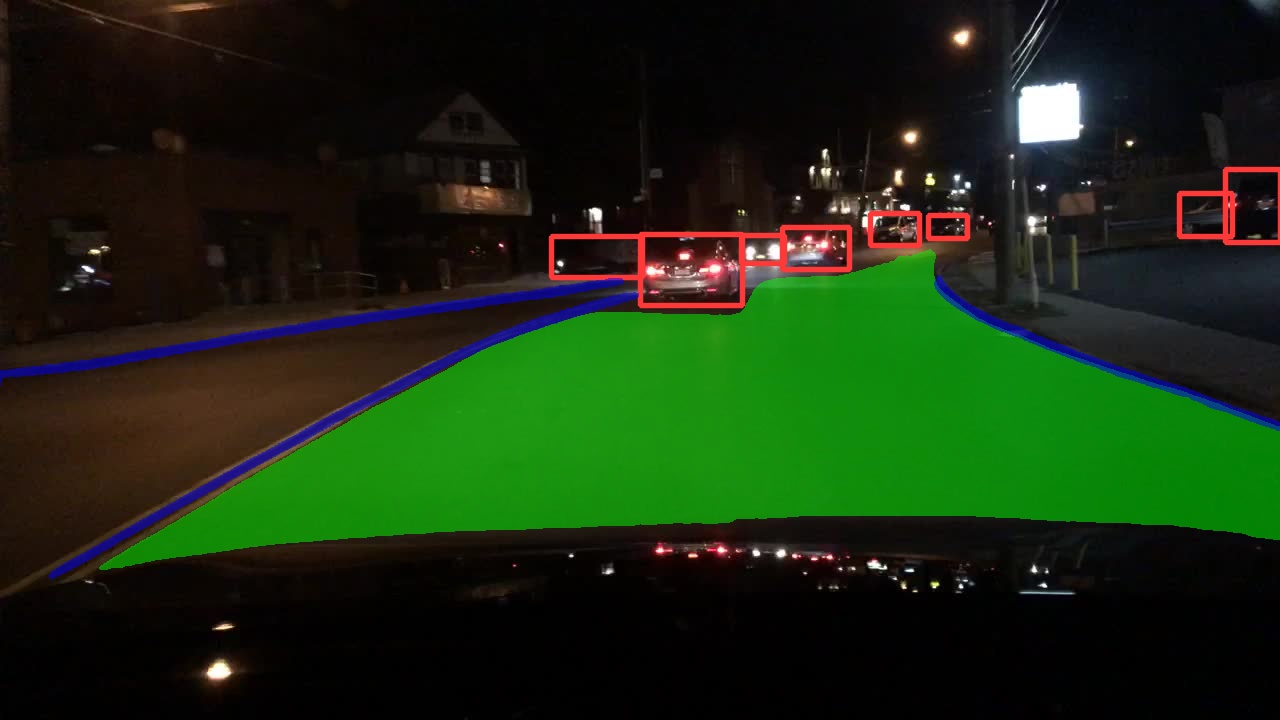}
    \end{subfigure}%
    \hspace{0.5cm}
    \begin{subfigure}{0.25\textwidth}
        \centering
        \includegraphics[width=\linewidth]{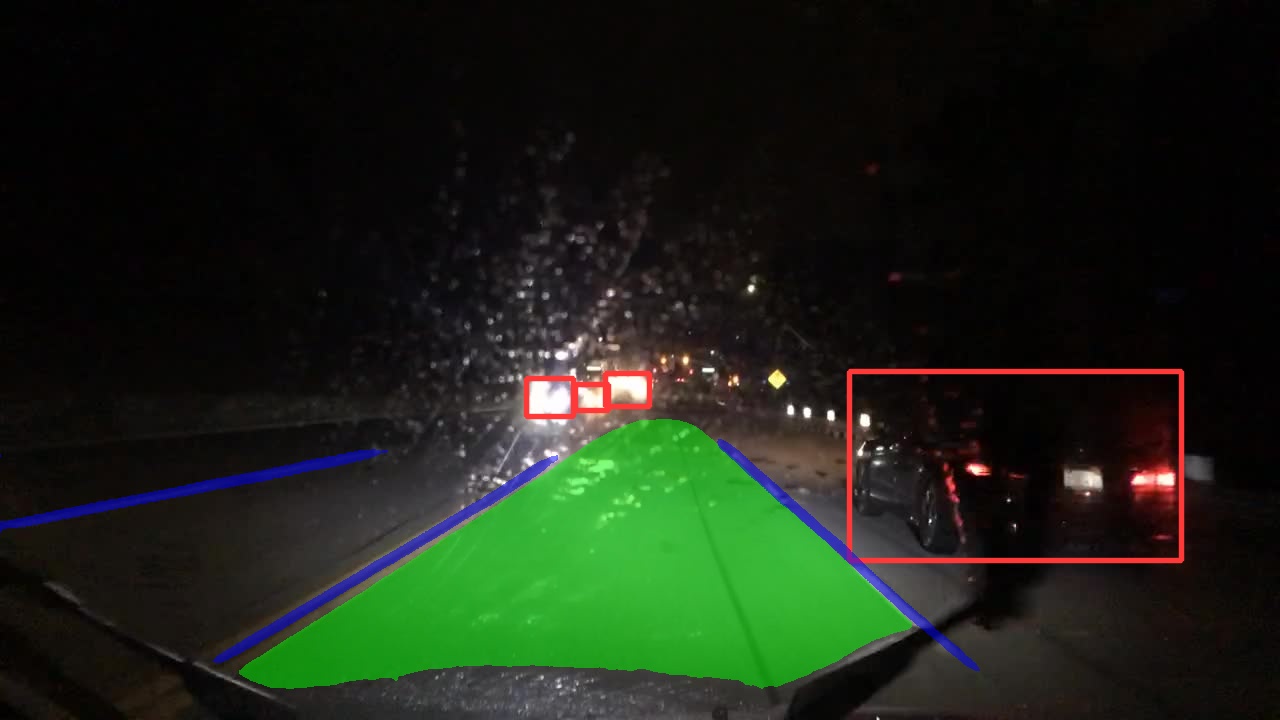}
    \end{subfigure}
    \caption{Visual Comparison of Results at Night}
    \label{fig:night_comparison}
\end{figure*}

Figure~\ref{fig:night_comparison} displays the results from night scenes. In nighttime scenes, the image quality decreases due to limited lighting and glare from oncoming vehicles. This is a challenge for the model to make accurate predictions. Under this challenge, our model consistently produces more accurate and smoother predictions for both lane lines and drivable areas. A-YOLOM(s) excels not only in the segmentation task but also in the detection task. Specifically, it surpasses A-YOLOM(n) by accurately detecting vehicles driving in the opposite direction, even amidst glare, and holds its own against YOLOP. Based on the detection results from the right image, YOLOP slightly outperforms our model in detecting distant vehicles at night. Nonetheless, our model produces significantly better segmentation results compared to YOLOP. Especially in the middle image, YOLOP mistakenly predicts the opposite lane as a drivable area. Such mispredictions are extremely hazardous for autonomous driving tasks. 

\begin{figure*}[!h]
    \centering
    
    \begin{subfigure}[b]{0.05\textwidth}
            \centering
            \rotatebox{90}{YOLOP}
            \vspace{0.6cm} 
    \end{subfigure}%
    \begin{subfigure}{0.25\textwidth}
        \centering
        \includegraphics[width=\linewidth]{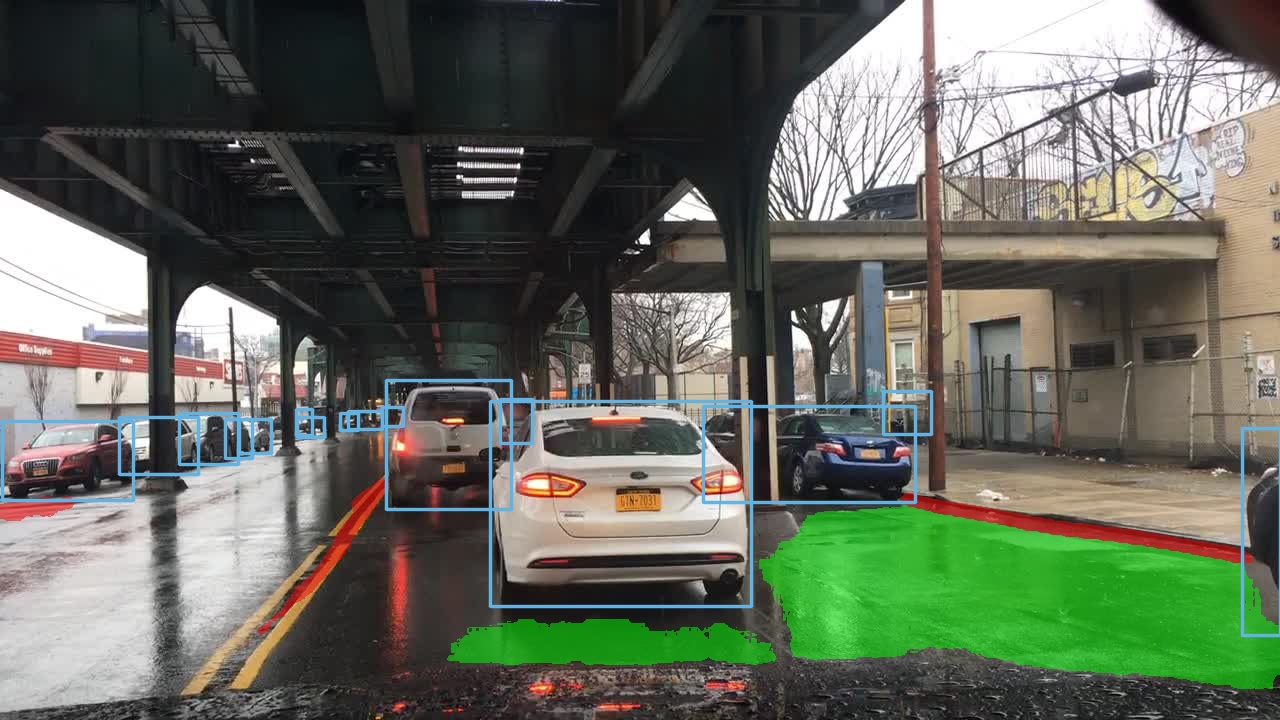}
    \end{subfigure}%
    \hspace{0.5cm}
    \begin{subfigure}{0.25\textwidth}
        \centering
        \includegraphics[width=\linewidth]{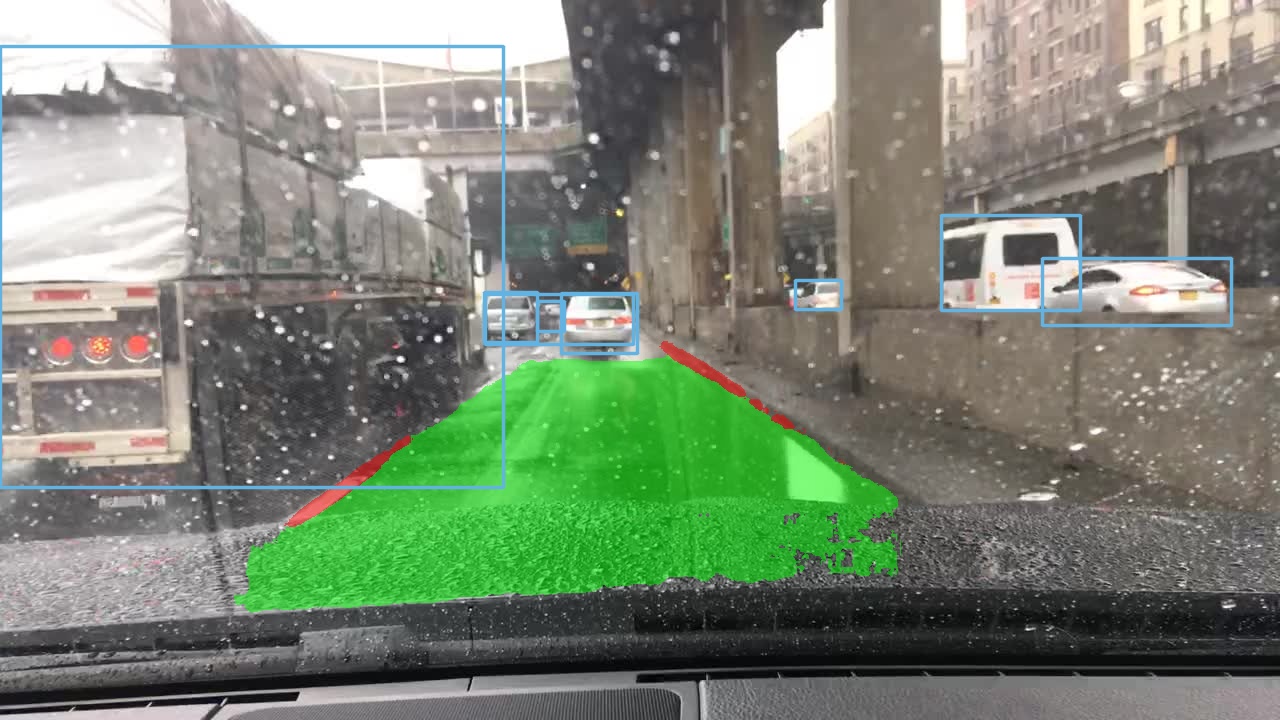}
    \end{subfigure}%
    \hspace{0.5cm}
    \begin{subfigure}{0.25\textwidth}
        \centering
        \includegraphics[width=\linewidth]{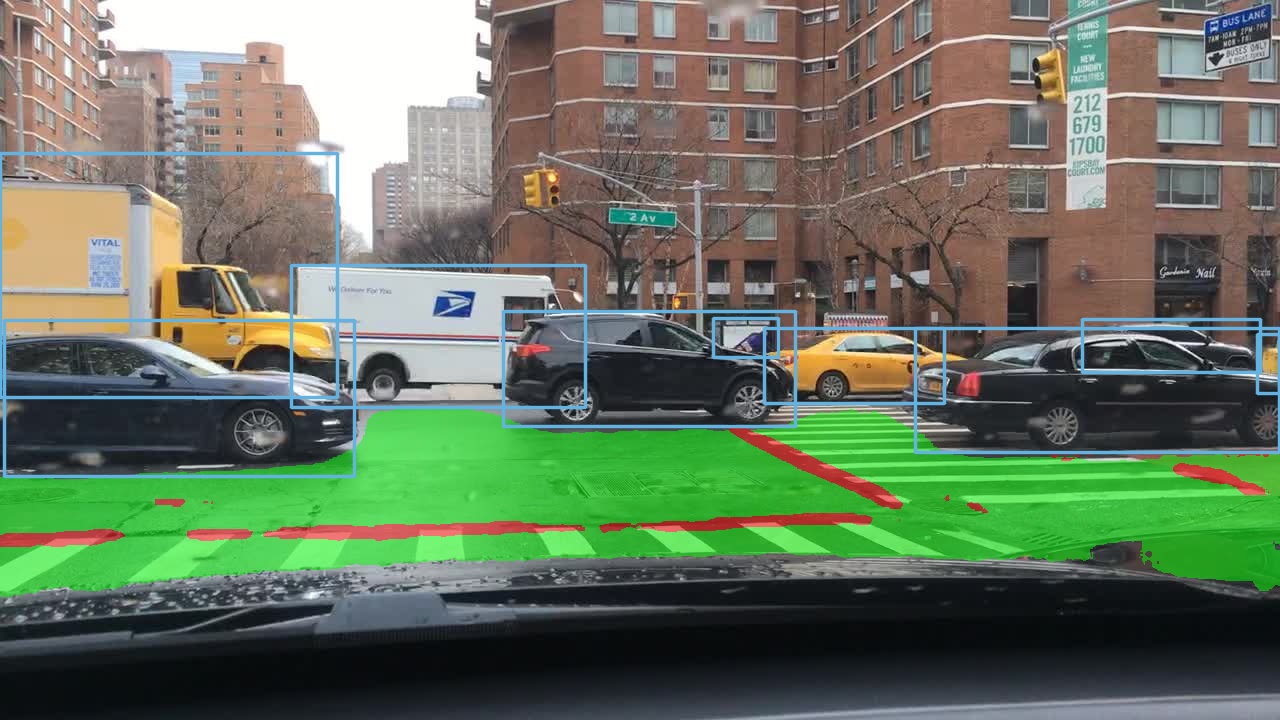}
    \end{subfigure}
    
    \medskip
    \begin{subfigure}[b]{0.05\textwidth}
        \centering
        \rotatebox{90}{A-YOLOM(n)}
        \vspace{0.2cm}
    \end{subfigure}%
    \begin{subfigure}{0.25\textwidth}
        \centering
        \includegraphics[width=\linewidth]{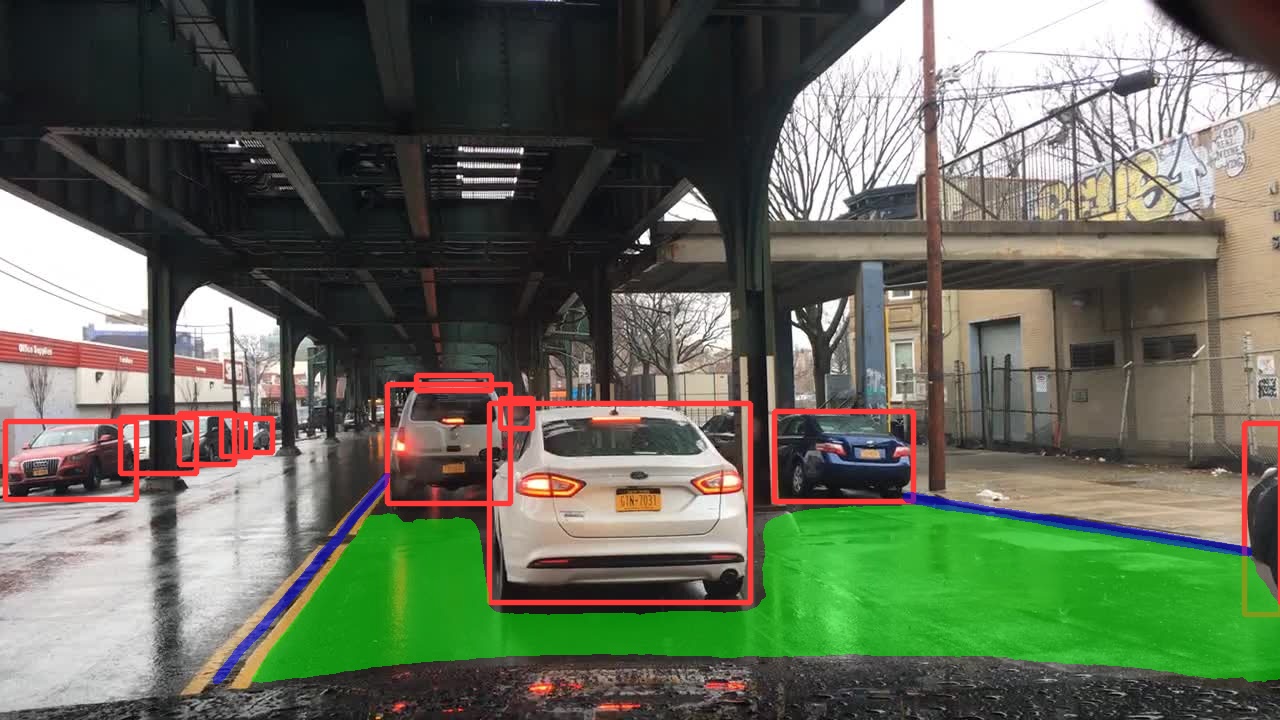}
    \end{subfigure}%
    \hspace{0.5cm}
    \begin{subfigure}{0.25\textwidth}
        \centering
        \includegraphics[width=\linewidth]{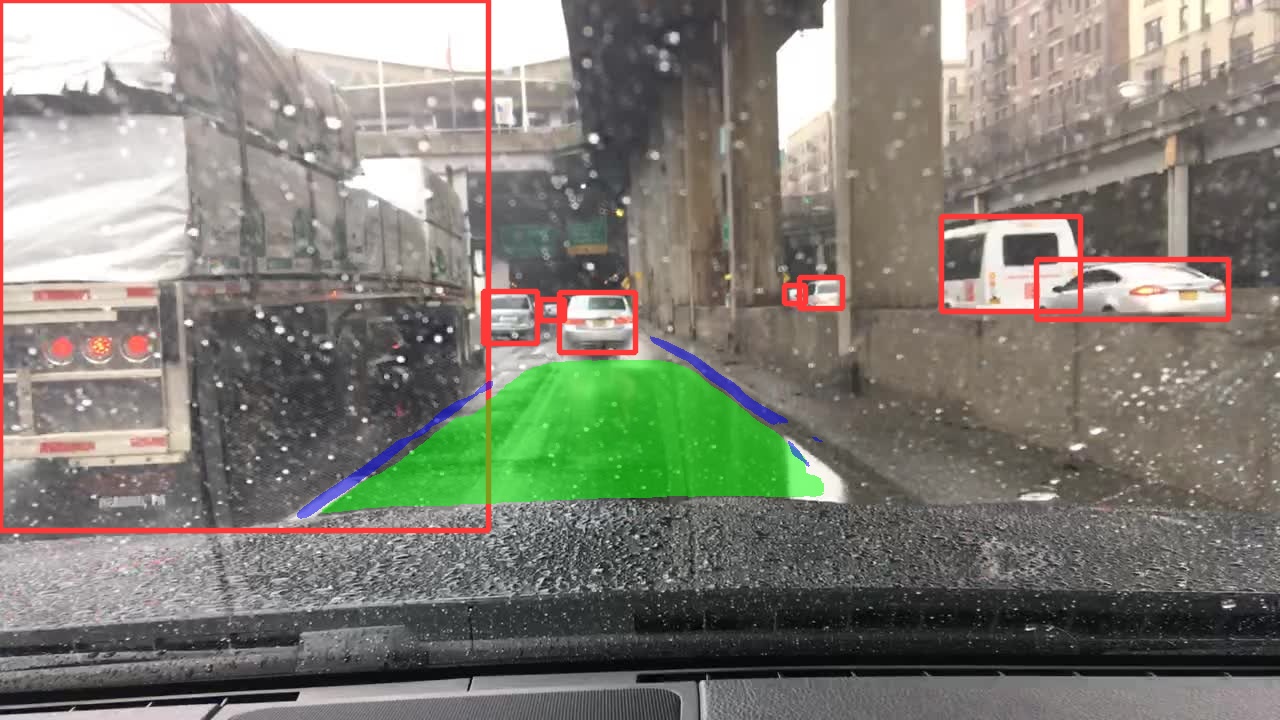}
    \end{subfigure}%
    \hspace{0.5cm}
    \begin{subfigure}{0.25\textwidth}
        \centering
        \includegraphics[width=\linewidth]{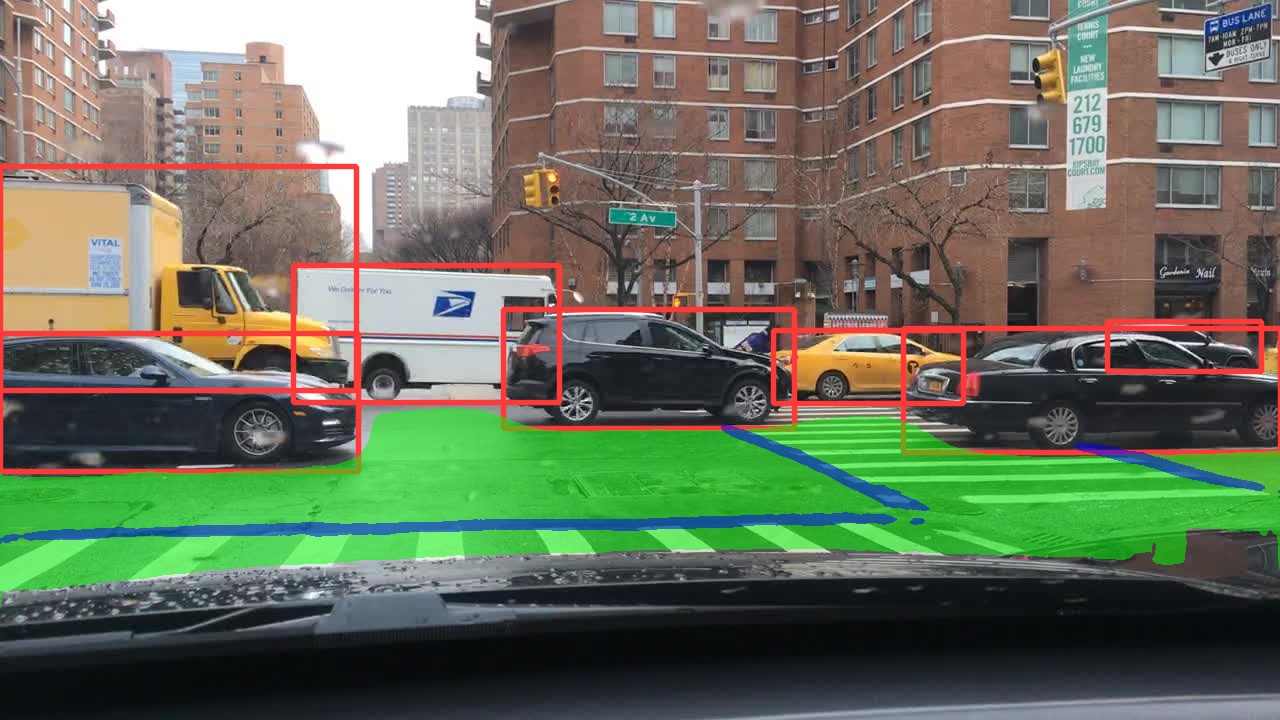}
    \end{subfigure}

    \medskip
    
    \begin{subfigure}[b]{0.05\textwidth}
        \centering
        \rotatebox{90}{ A-YOLOM(s)}
        \vspace{0.1cm}
    \end{subfigure}%
    \begin{subfigure}{0.25\textwidth}
        \centering
        \includegraphics[width=\linewidth]{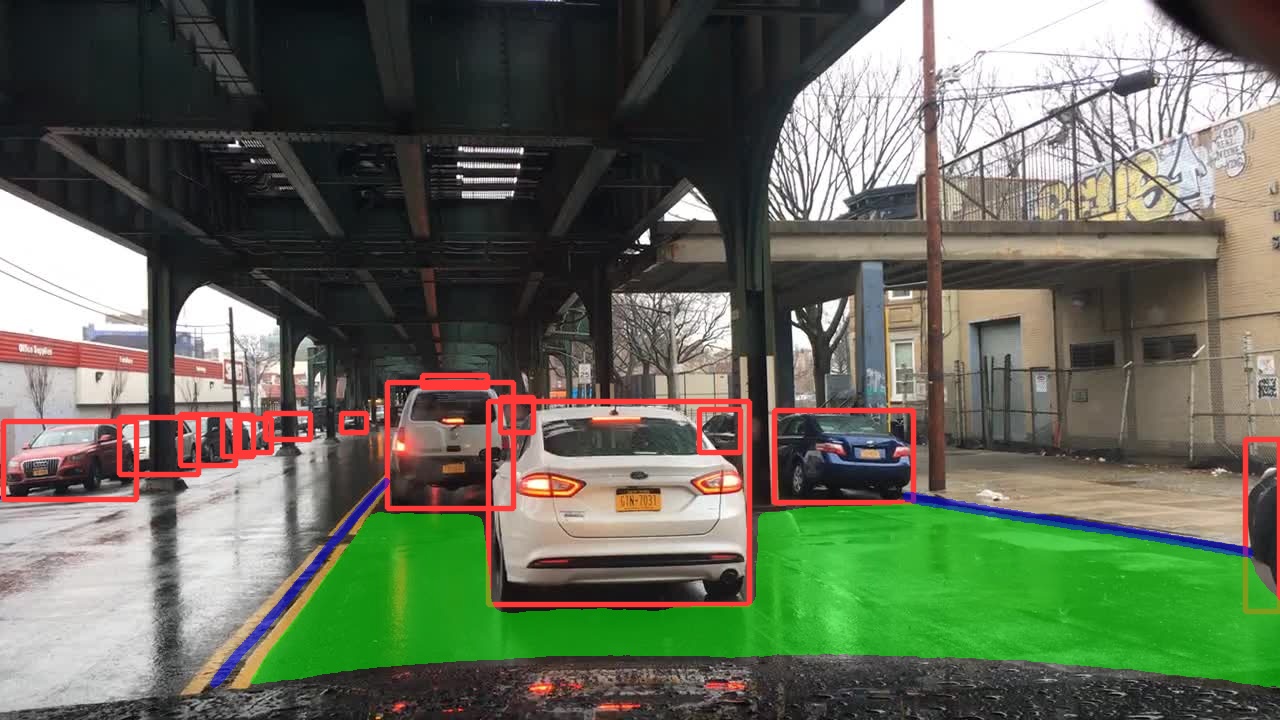}
    \end{subfigure}%
    \hspace{0.5cm}
    \begin{subfigure}{0.25\textwidth}
        \centering
        \includegraphics[width=\linewidth]{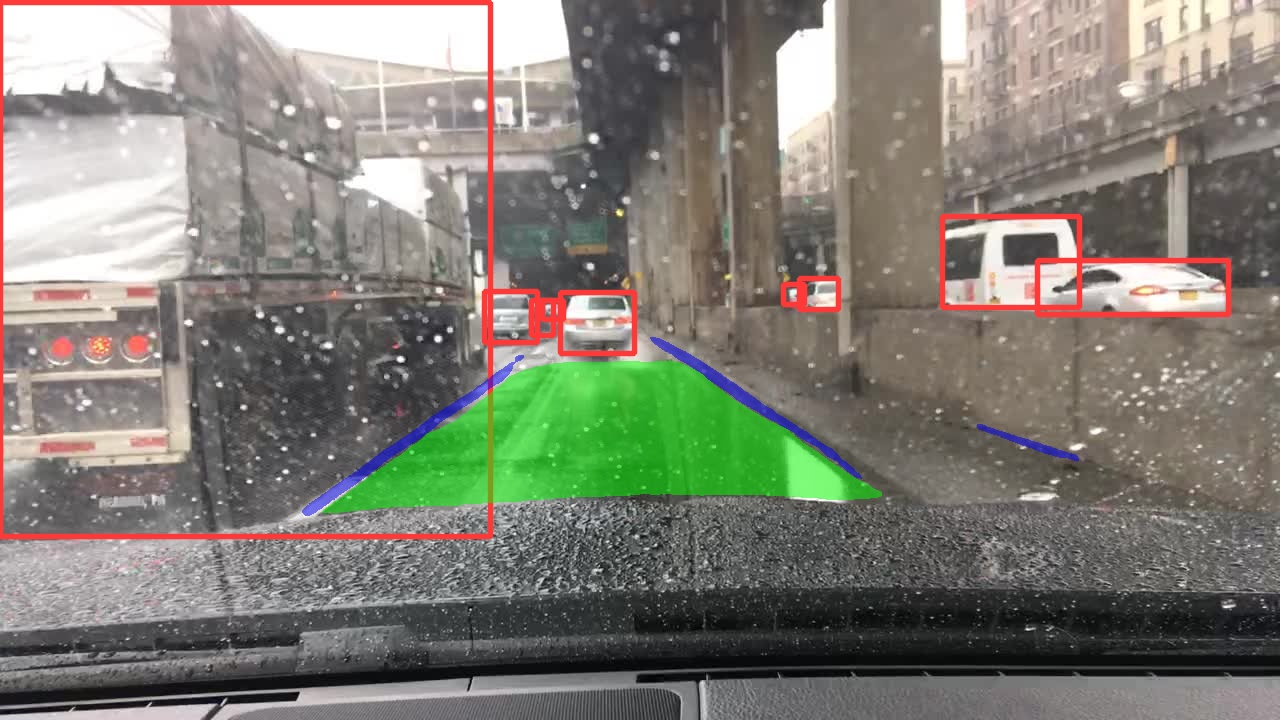}
    \end{subfigure}%
    \hspace{0.5cm}
    \begin{subfigure}{0.25\textwidth}
        \centering
        \includegraphics[width=\linewidth]{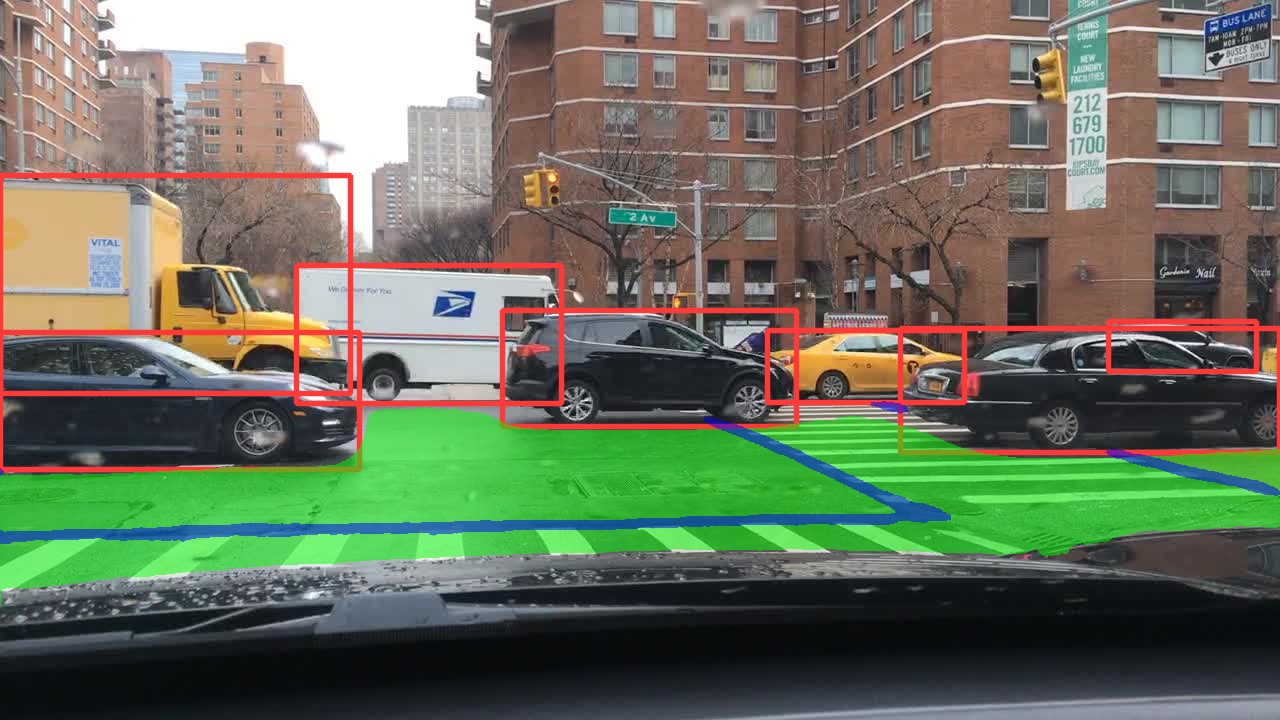}
    \end{subfigure}
    \caption{Visual Comparison of Results on Rainy Day}
    \label{fig:rainy_day_comparison}
\end{figure*}

\begin{figure*}[!h]
    \centering
    \begin{subfigure}[b]{0.05\textwidth}
        \centering
        \rotatebox{90}{YOLOP}
        \vspace{0.6cm} 
    \end{subfigure}%
    \begin{subfigure}{0.25\textwidth}
        \centering
        \includegraphics[width=\linewidth]{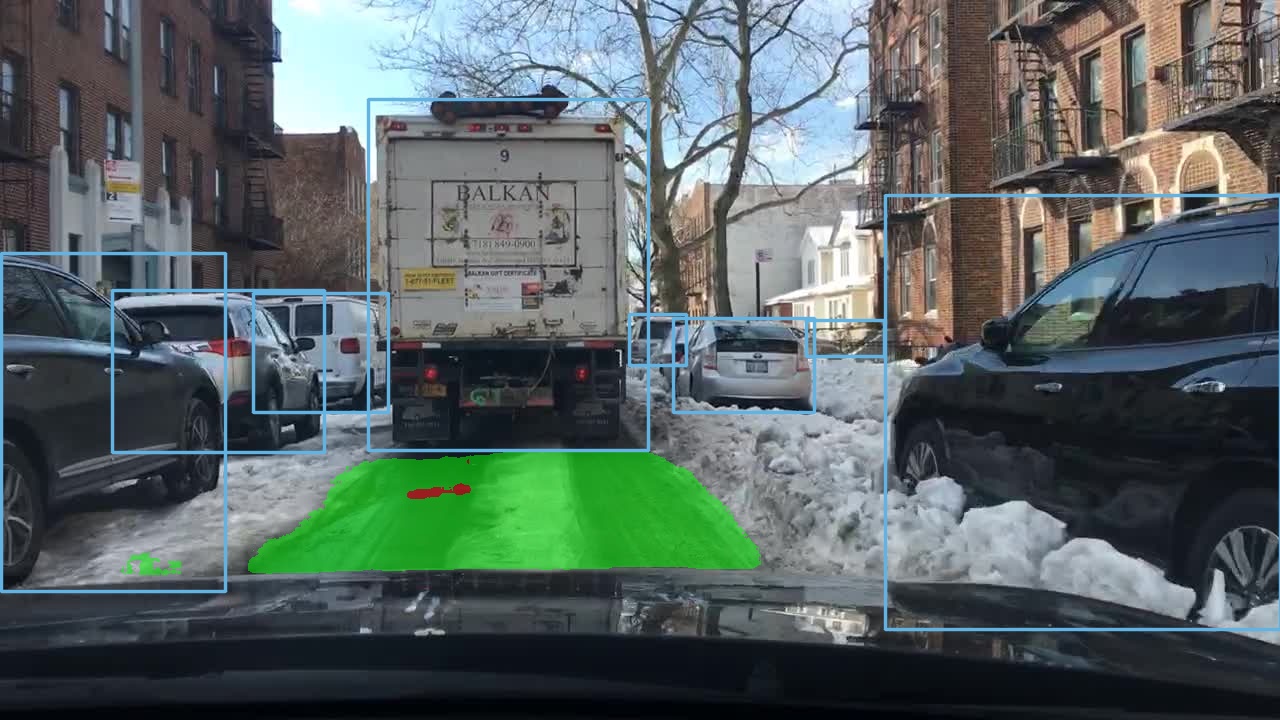}
    \end{subfigure}%
    \hspace{0.5cm}
    \begin{subfigure}{0.25\textwidth}
        \centering
        \includegraphics[width=\linewidth]{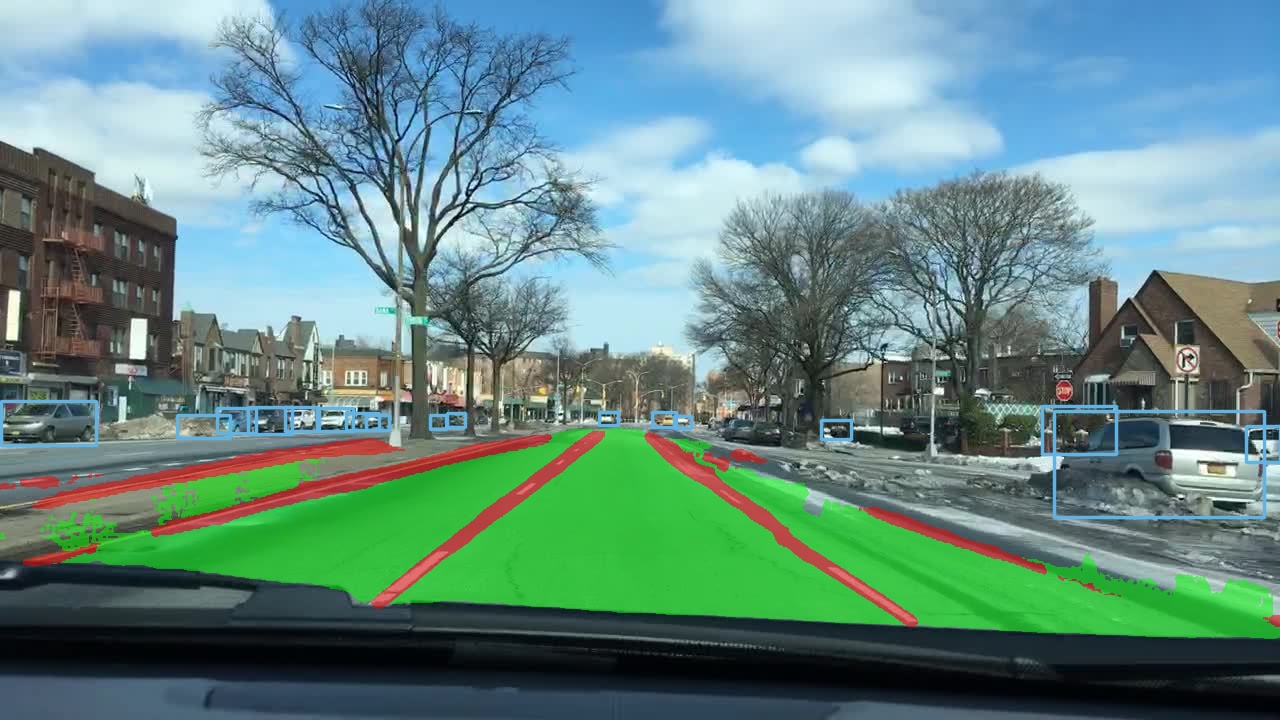}
    \end{subfigure}%
    \hspace{0.5cm}
    \begin{subfigure}{0.25\textwidth}
        \centering
        \includegraphics[width=\linewidth]{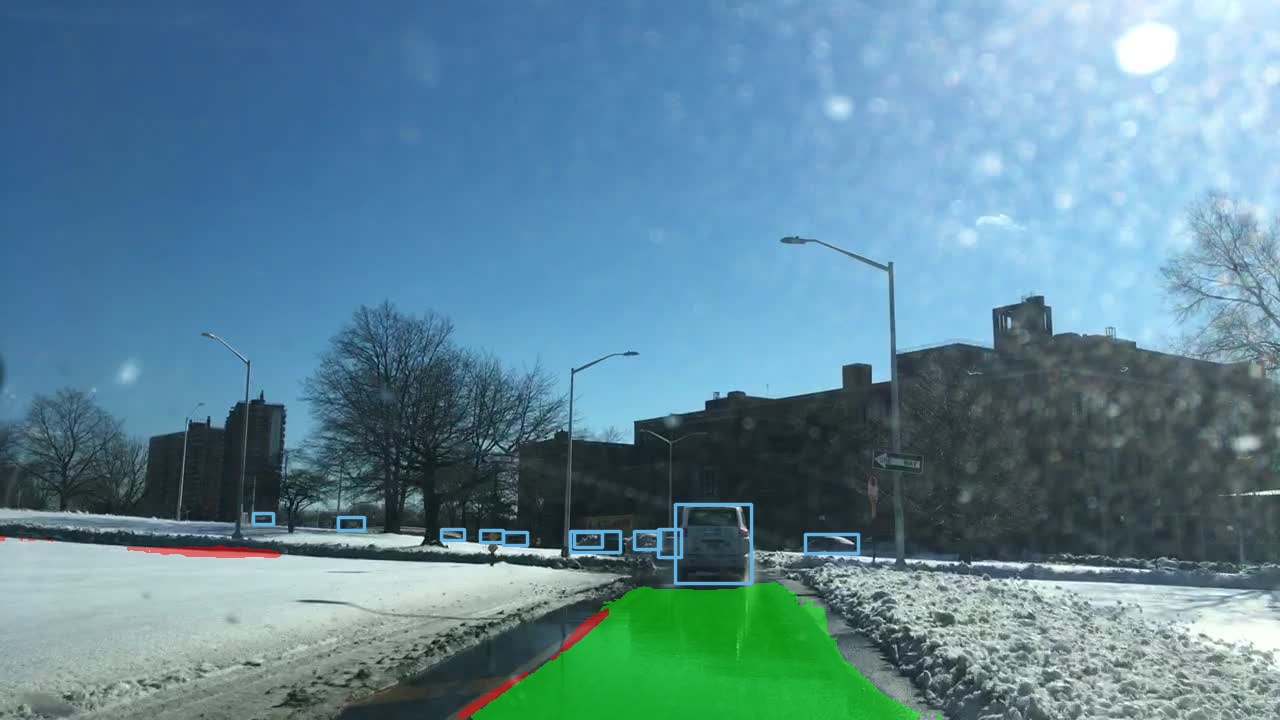}
    \end{subfigure}
    
    \medskip

    \begin{subfigure}[b]{0.05\textwidth}
        \centering
        \rotatebox{90}{A-YOLOM(n)}
        \vspace{0.2cm}
    \end{subfigure}%
    \begin{subfigure}{0.25\textwidth}
        \centering
        \includegraphics[width=\linewidth]{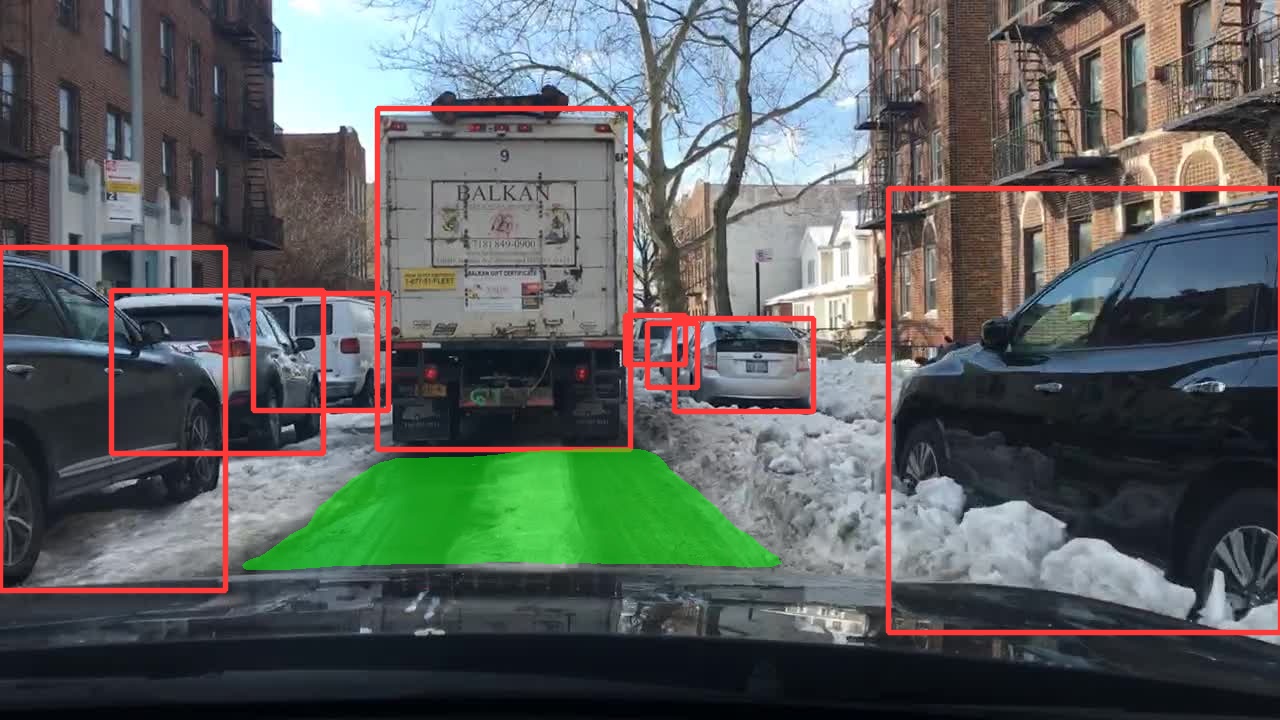}
    \end{subfigure}%
    \hspace{0.5cm}
    \begin{subfigure}{0.25\textwidth}
        \centering
        \includegraphics[width=\linewidth]{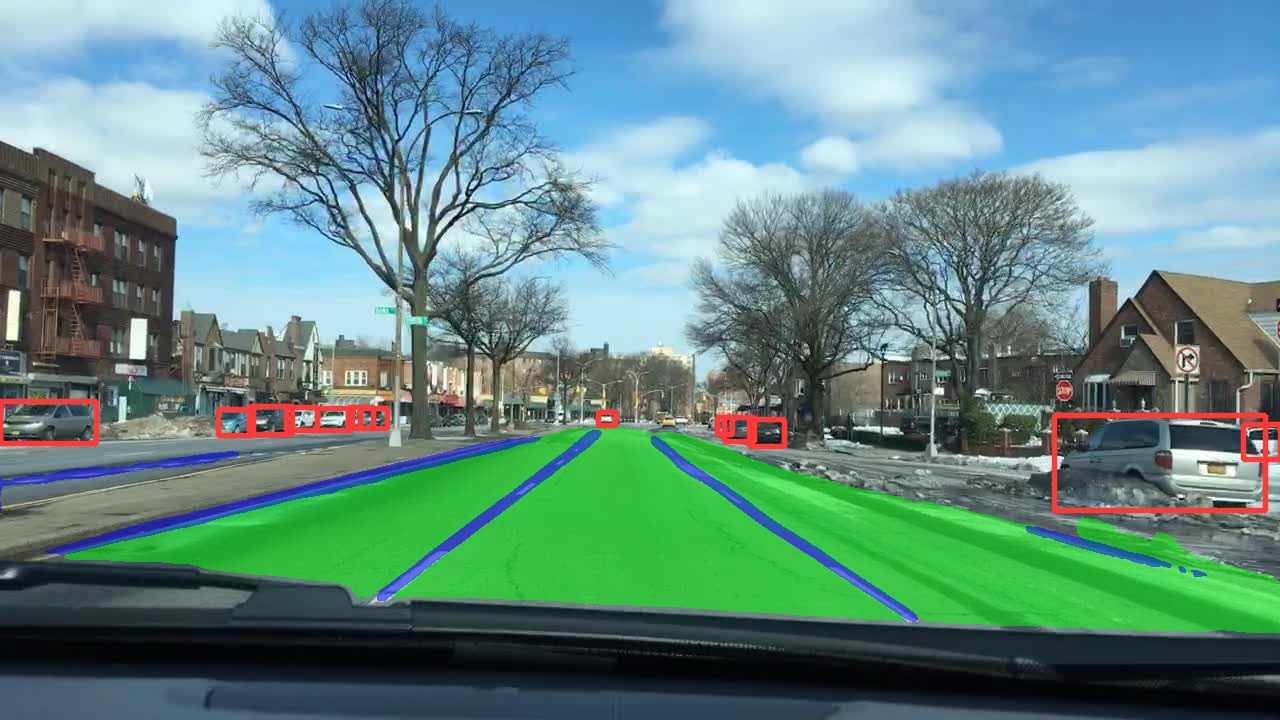}
    \end{subfigure}%
    \hspace{0.5cm}
    \begin{subfigure}{0.25\textwidth}
        \centering
        \includegraphics[width=\linewidth]{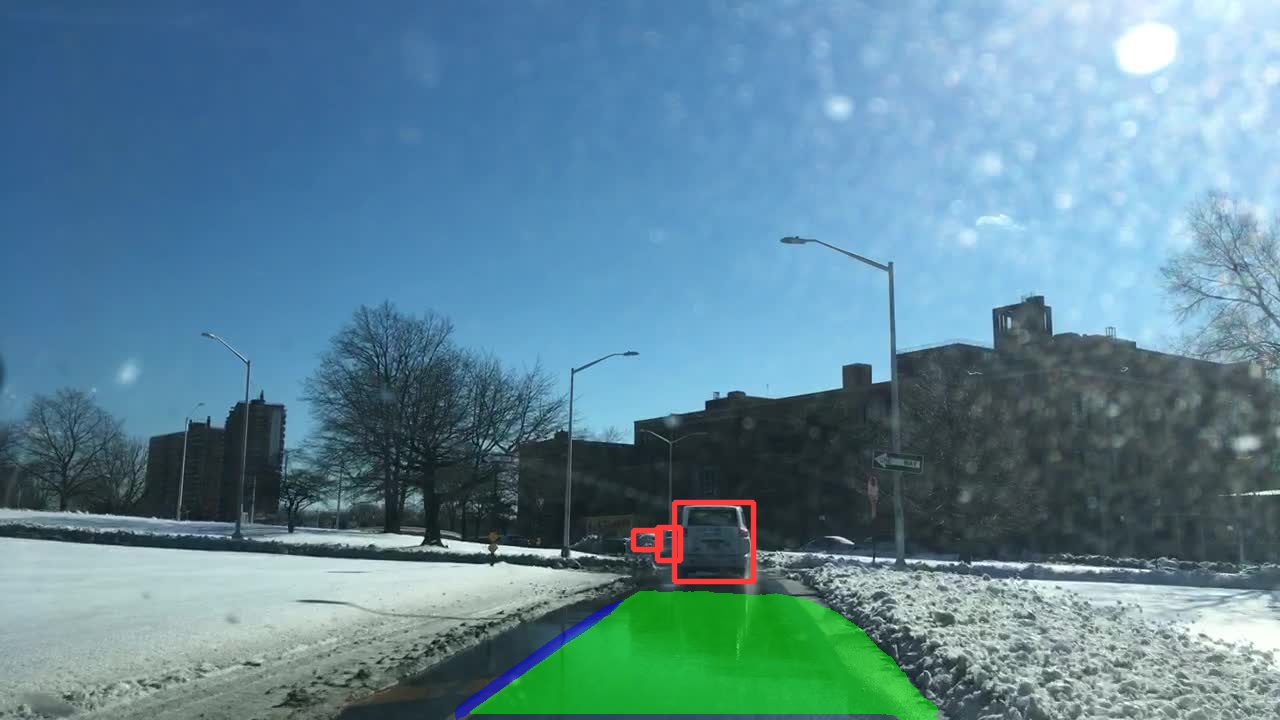}
    \end{subfigure}

    \medskip
    
    \begin{subfigure}[b]{0.05\textwidth}
        \centering
        \rotatebox{90}{ A-YOLOM(s)}
        \vspace{0.1cm}
    \end{subfigure}%
    \begin{subfigure}{0.25\textwidth}
        \centering
        \includegraphics[width=\linewidth]{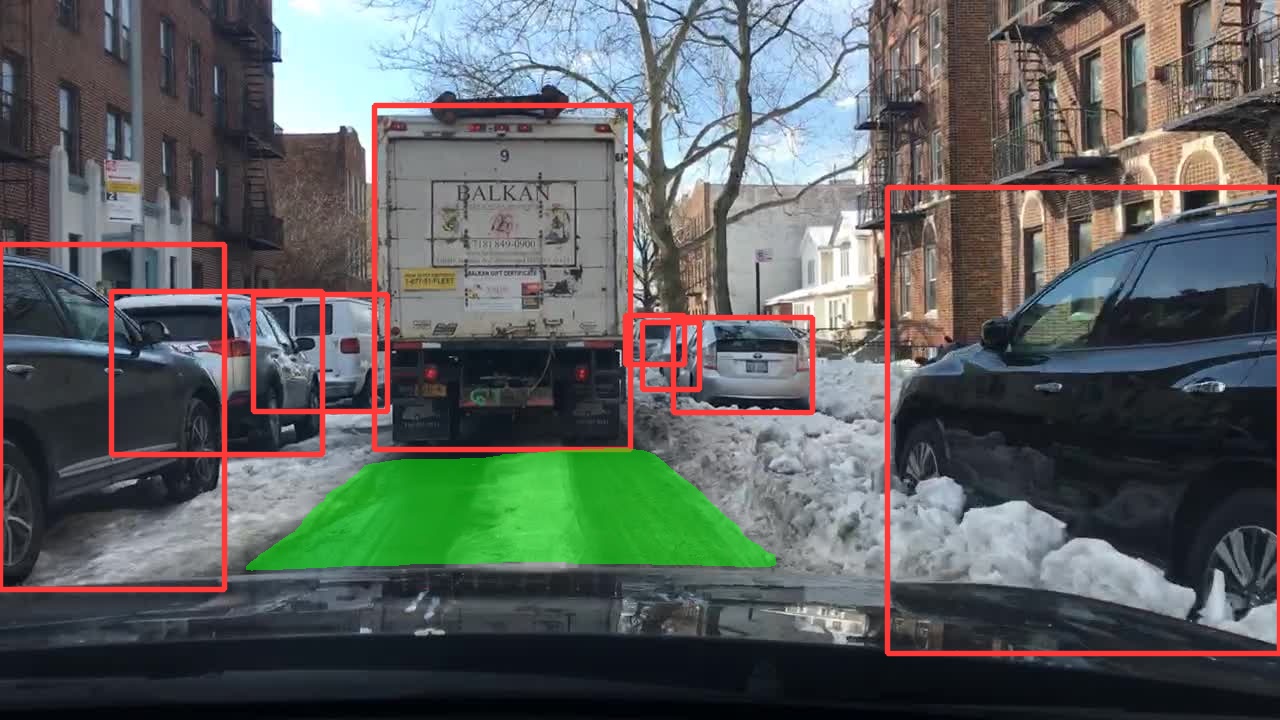}
    \end{subfigure}%
    \hspace{0.5cm}
    \begin{subfigure}{0.25\textwidth}
        \centering
        \includegraphics[width=\linewidth]{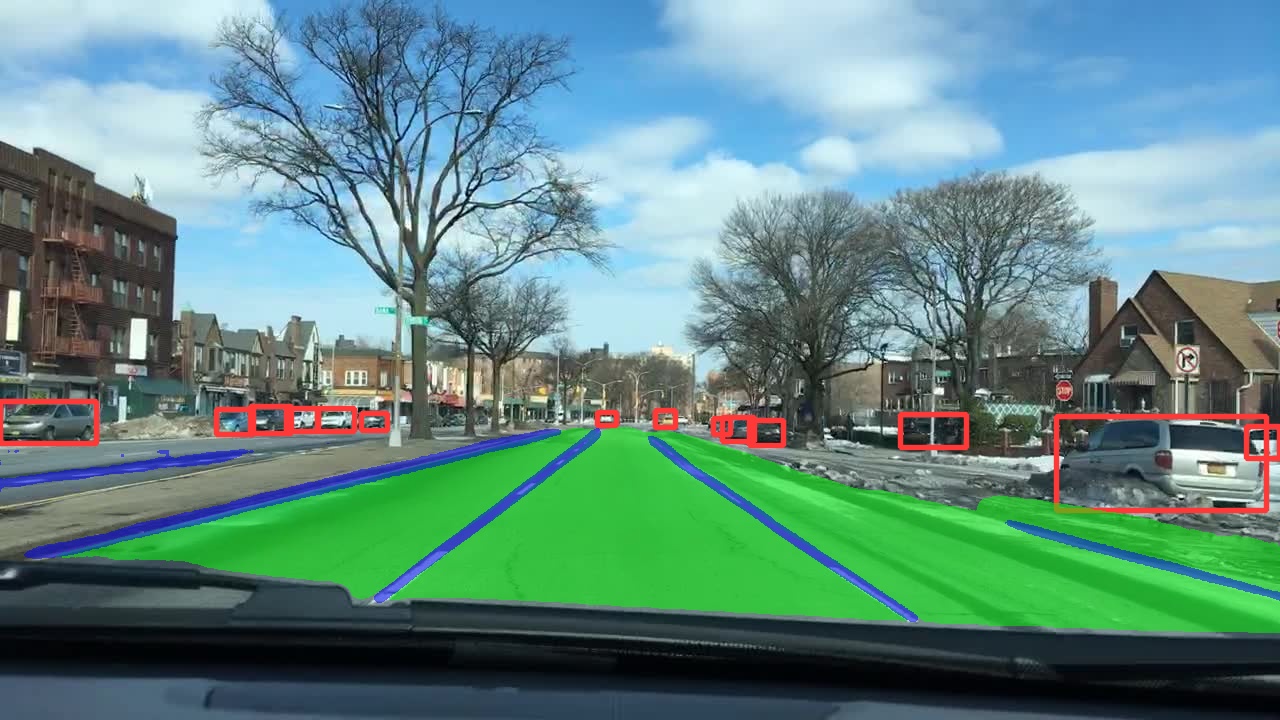}
    \end{subfigure}%
    \hspace{0.5cm}
    \begin{subfigure}{0.25\textwidth}
        \centering
        \includegraphics[width=\linewidth]{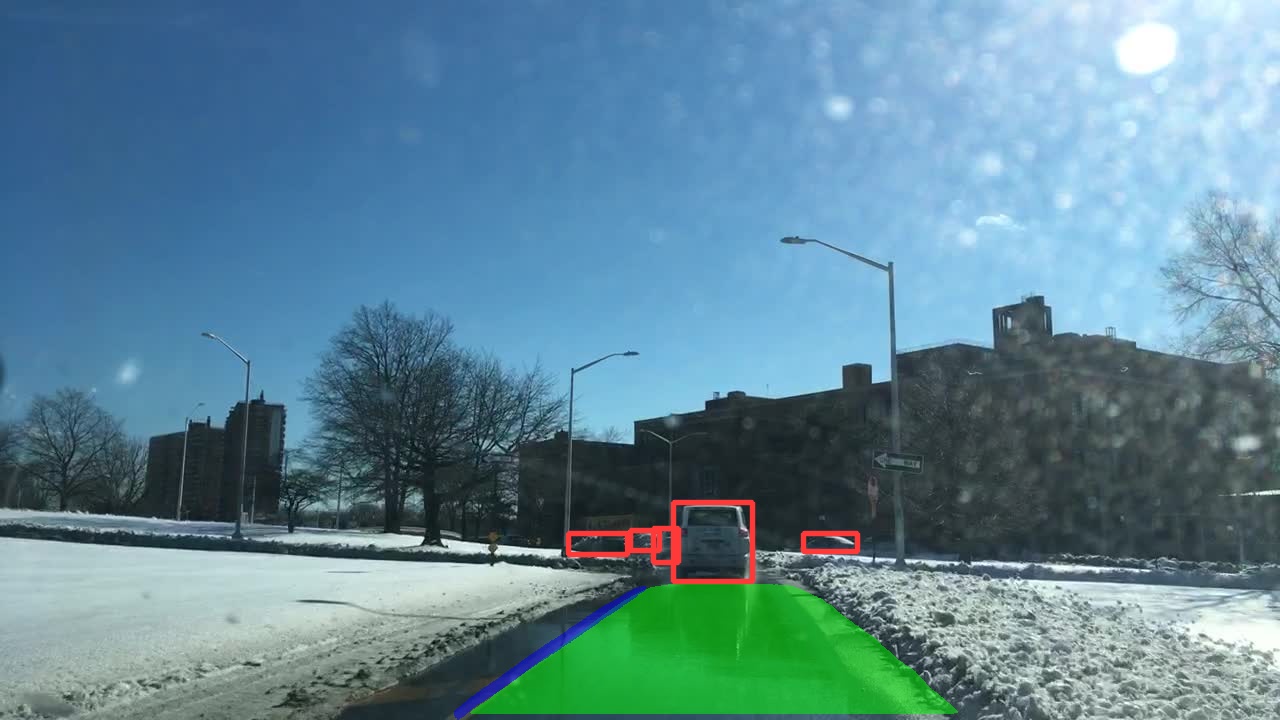}
    \end{subfigure}
    \caption{Visual Comparison of Results on Snow Day}
    \label{fig:snow_day_comparison}
\end{figure*}

Figure~\ref{fig:rainy_day_comparison} displays the results from a rainy day. Rain will increase road surface reflection, thereby impacting the driver's judgment. This issue also manifests in deep learning models. Moreover, raindrops on the windshield, due to their refractive and scattering effects, can blur the entire image, creating considerable challenges for the model's predictions. In the results from the left image, we can observe that YOLOP cannot accurately segment the drivable area due to reflections on the road surface. A similar situation is evident in the middle image. YOLOP even struggles to distinguish between the car hood and the road. However, our model still delivers outstanding performance in this scenario. Especially for A-YOLOM(s), it is less affected by the reflections on the road surface. Our model also demonstrates superior lane line detection compared to YOLOP, as evidenced by the right image. However, in this scene, YOLOP outperforms us in detection tasks. As observed in the left image, YOLOP can detect more vehicles at a farther distance.

Figure~\ref{fig:snow_day_comparison} displays the results from a snow day. In snowy conditions, accumulated snow can obscure the road or lanes, posing additional challenges for models. Some accumulated snow is cleared into snow mounds on the side of the road, which poses challenges for the model's vehicle prediction. For instance, on the left in the middle image, YOLOP mistakenly detects a snow mound as a vehicle. If there's a small snow mound on the road, vehicles can typically pass over it. Misidentifying such little snow mounds as stationary vehicles could be dangerous. Therefore, a higher recall is not always better. It's crucial to strike a balance between recall and precision. In all three results presented, our model significantly outperforms YOLOP in the segmentation task, with A-YOLOM(s) being particularly noteworthy. However, based on the results from the right image, our model lags behind YOLOP in detecting distant vehicles, particularly when using the A-YOLOM(n).

In various weather conditions, our model consistently delivers more accurate and smoother segmentation results. However, our model's ability to detect distant and small targets falls slightly compared to YOLOP in adverse weather conditions.

\subsection{Ablation Studies}
\label{subsec: Ablation Studies}
\begin{figure*}[!h]
    \centering
    
    \begin{subfigure}[b]{0.05\textwidth}
        \centering
        \rotatebox{90}{YOLOP}
        \vspace{0.6cm} 
    \end{subfigure}%
    \begin{subfigure}{0.25\textwidth}
        \centering
        \includegraphics[width=\linewidth]{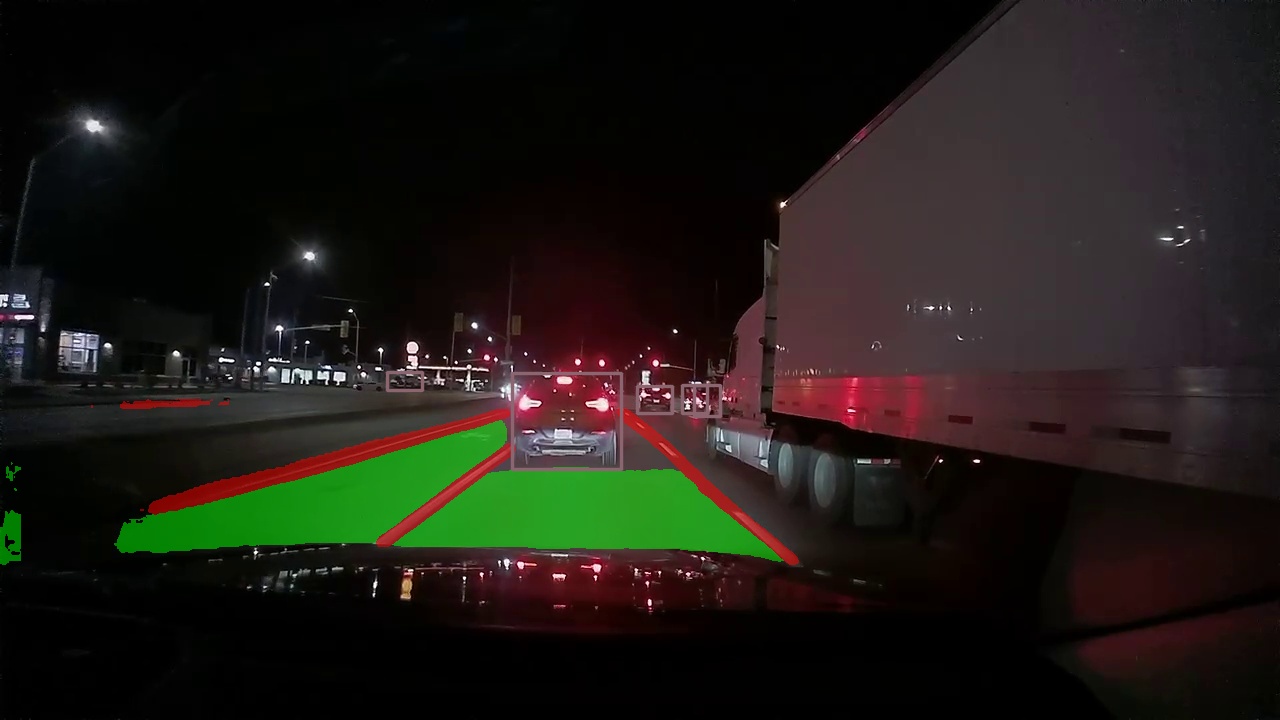}
    \end{subfigure}%
    \hspace{0.5cm}
    \begin{subfigure}{0.25\textwidth}
        \centering
        \includegraphics[width=\linewidth]{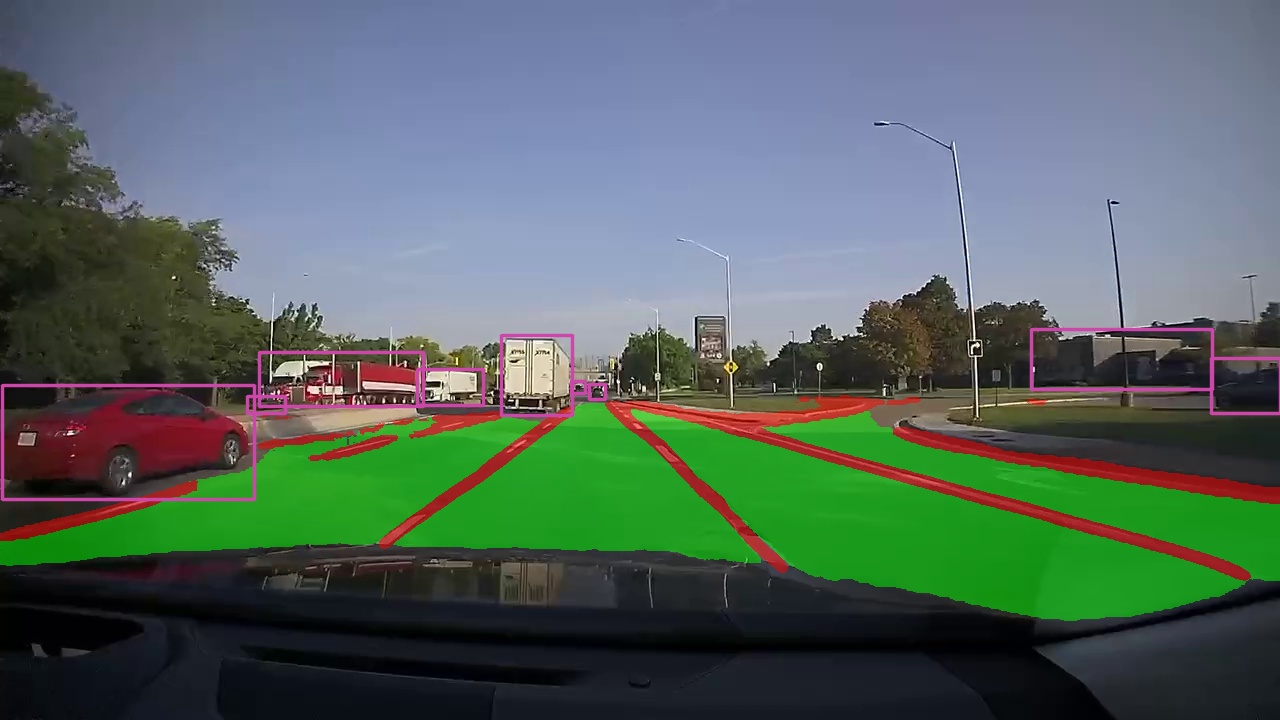}
    \end{subfigure}%
    \hspace{0.5cm}
    \begin{subfigure}{0.25\textwidth}
        \centering
        \includegraphics[width=\linewidth]{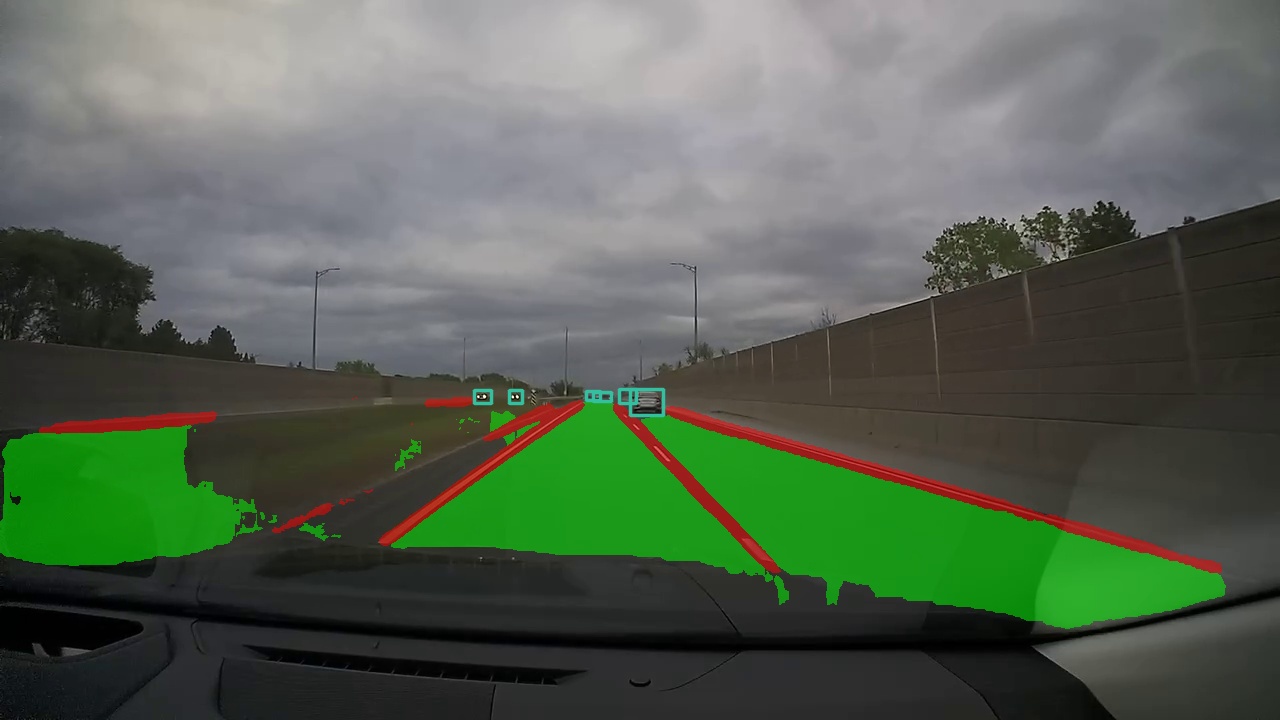}
    \end{subfigure}
    
    \medskip
    
    \begin{subfigure}[b]{0.05\textwidth}
        \centering
        \rotatebox{90}{A-YOLOM(n)}
        \vspace{0.2cm}
    \end{subfigure}%
    \begin{subfigure}{0.25\textwidth}
        \centering
        \includegraphics[width=\linewidth]{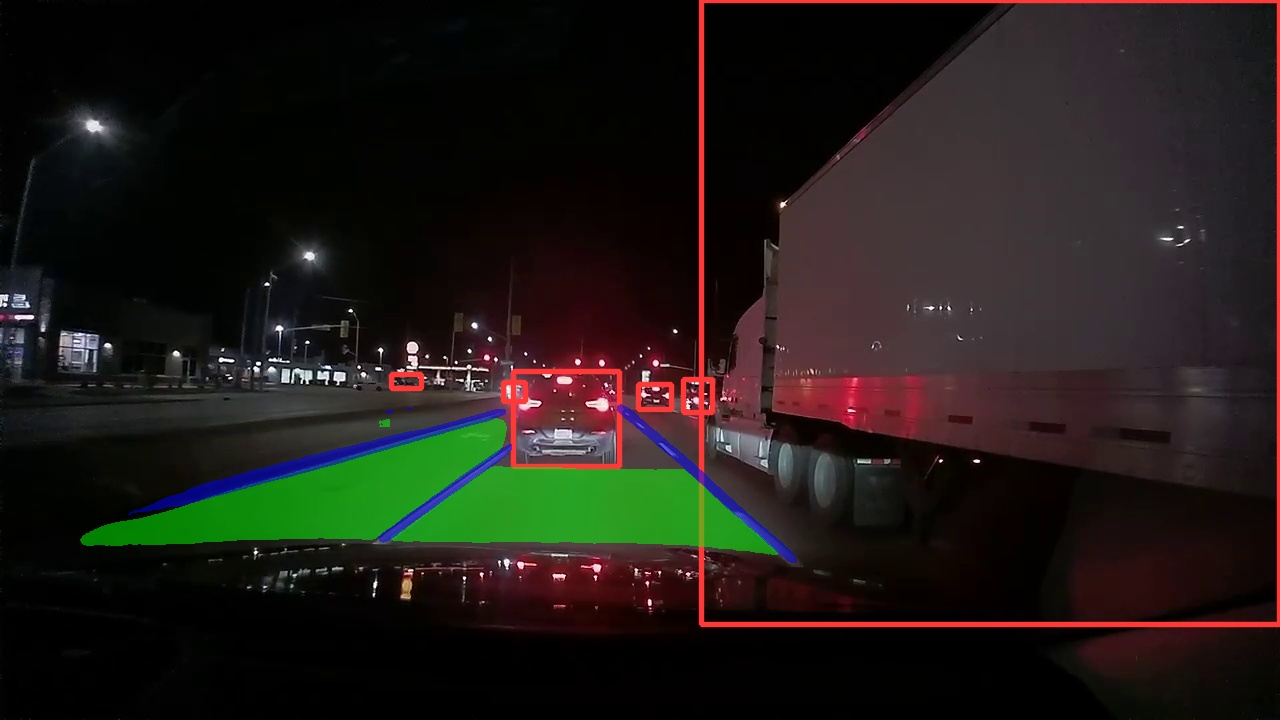}
    \end{subfigure}%
    \hspace{0.5cm}
    \begin{subfigure}{0.25\textwidth}
        \centering
        \includegraphics[width=\linewidth]{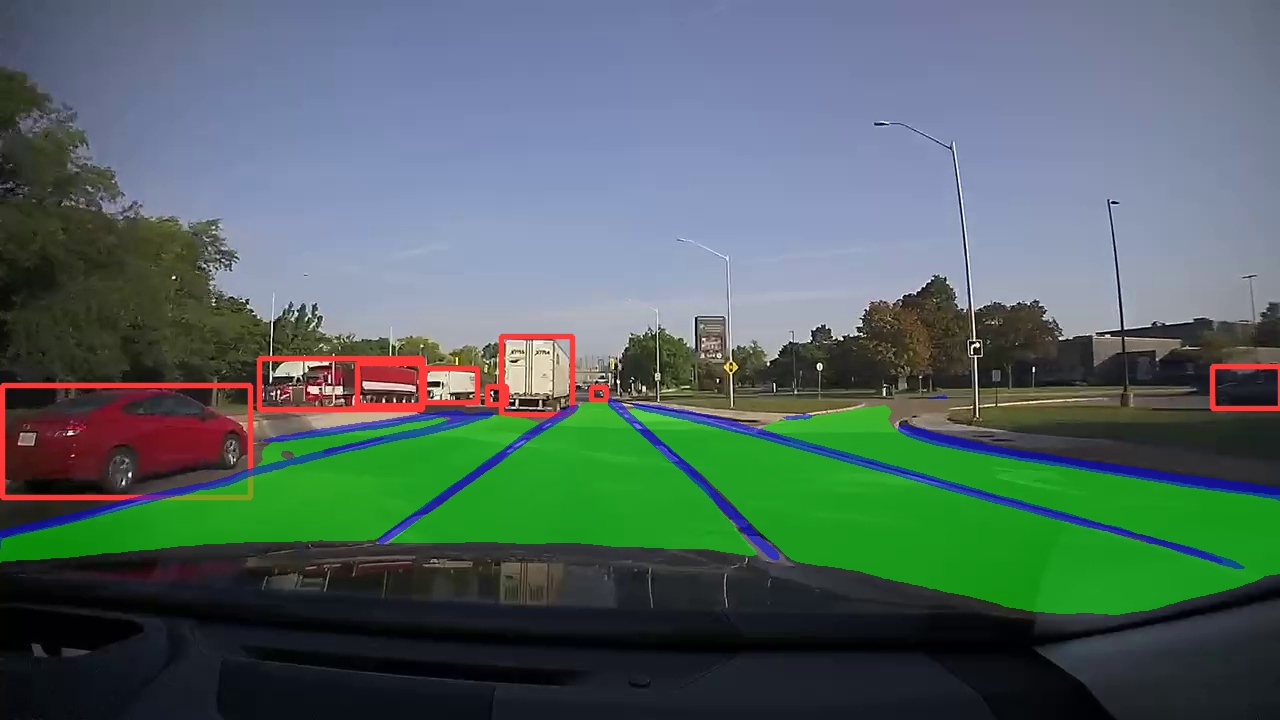}
    \end{subfigure}%
    \hspace{0.5cm}
    \begin{subfigure}{0.25\textwidth}
        \centering
        \includegraphics[width=\linewidth]{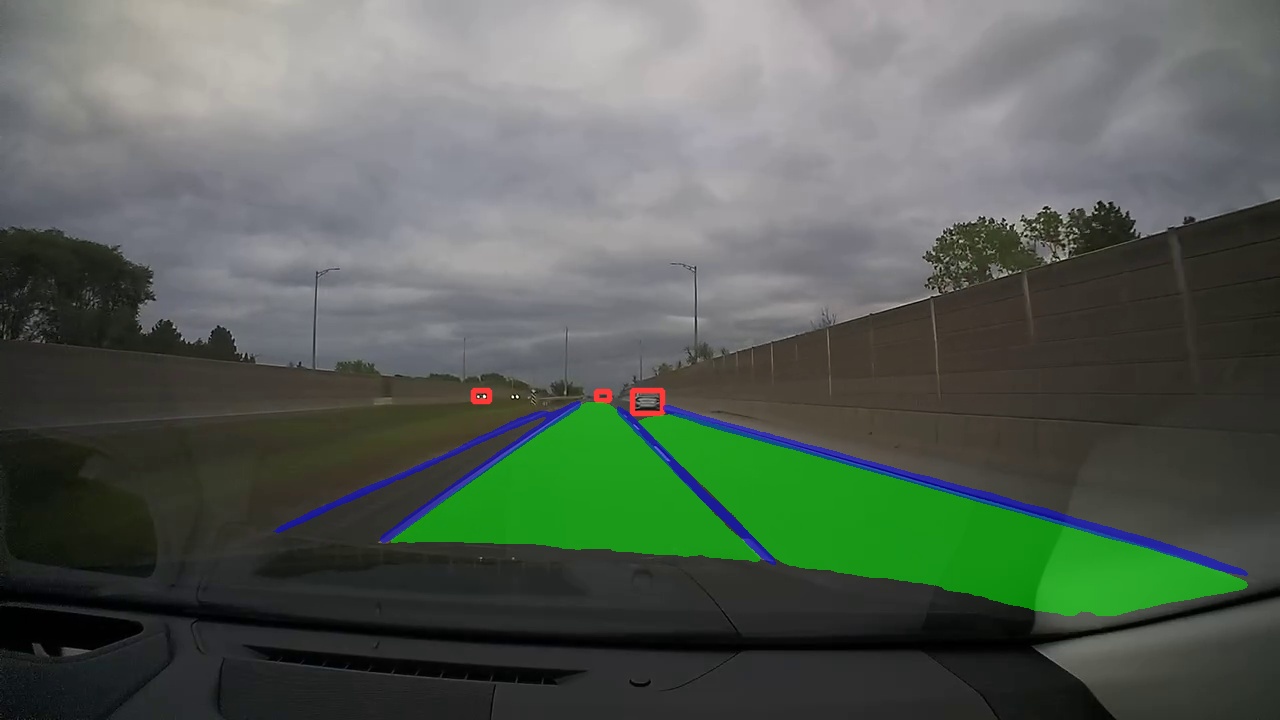}
    \end{subfigure}

    \medskip
    
    \begin{subfigure}[b]{0.05\textwidth}
        \centering
        \rotatebox{90}{ A-YOLOM(s)}
        \vspace{0.1cm}
    \end{subfigure}%
    \begin{subfigure}{0.25\textwidth}
        \centering
        \includegraphics[width=\linewidth]{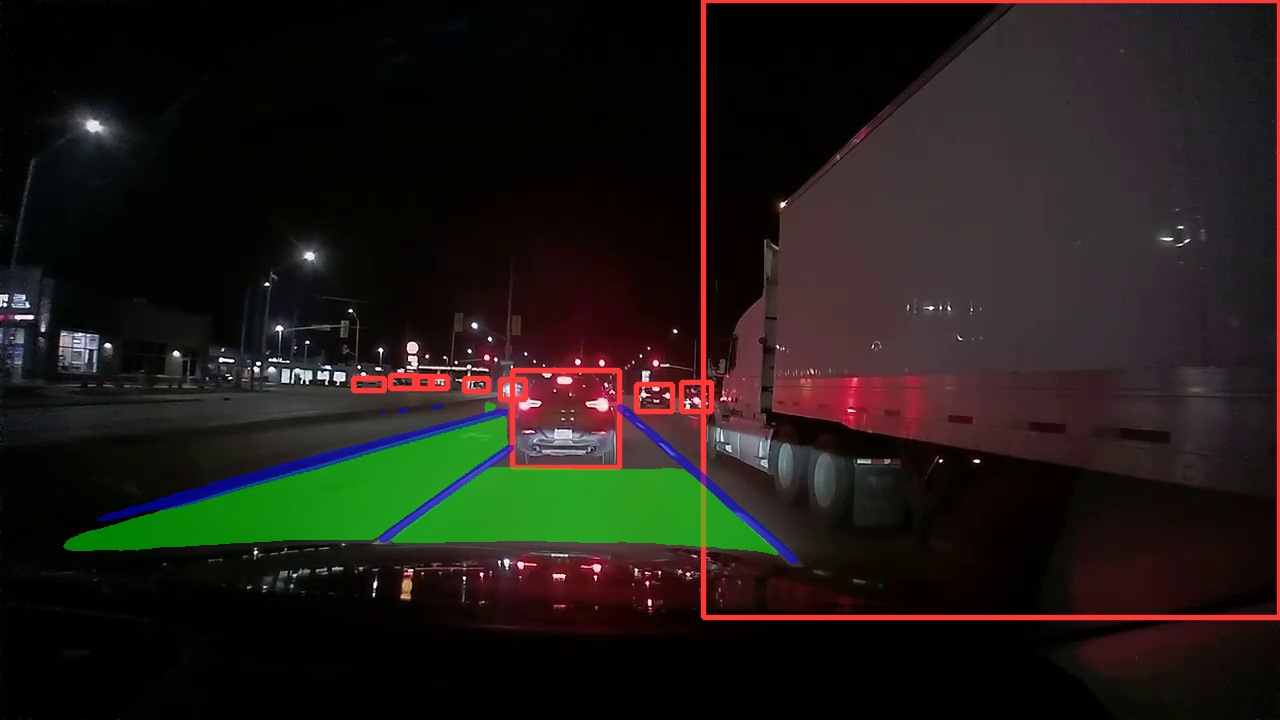}
    \end{subfigure}%
    \hspace{0.5cm}
    \begin{subfigure}{0.25\textwidth}
        \centering
        \includegraphics[width=\linewidth]{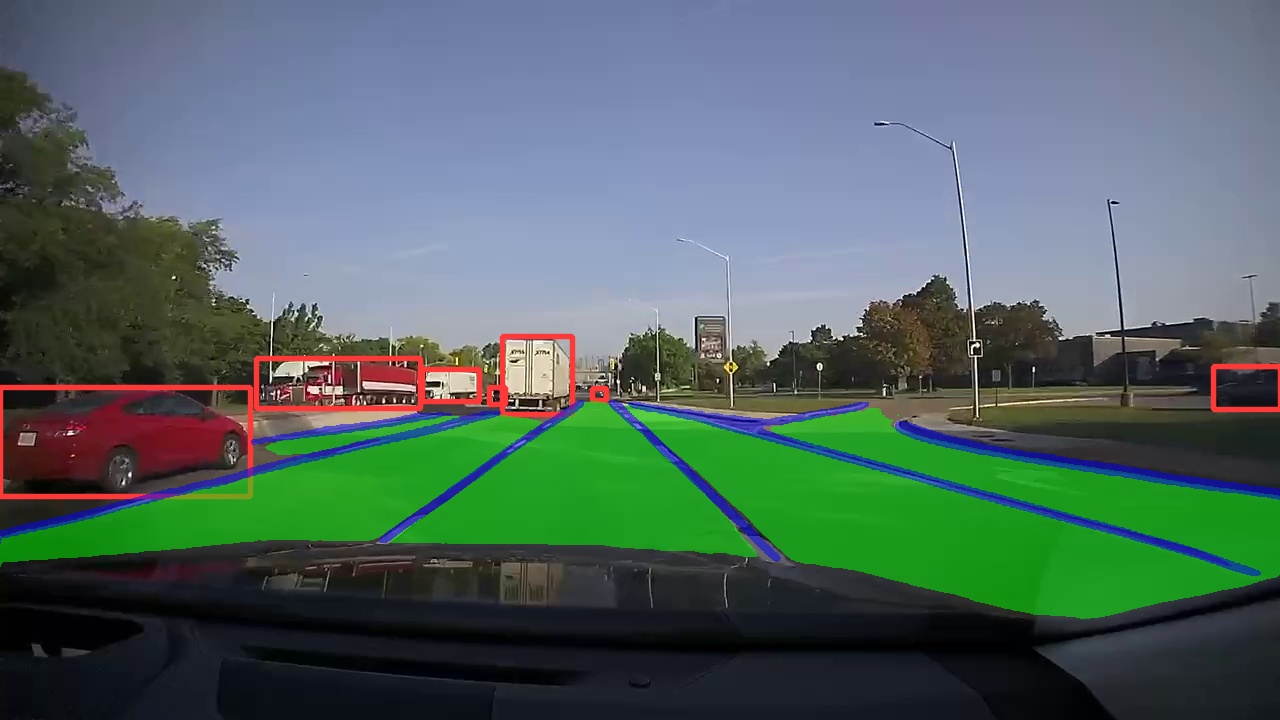}
    \end{subfigure}%
    \hspace{0.5cm}
    \begin{subfigure}{0.25\textwidth}
        \centering
        \includegraphics[width=\linewidth]{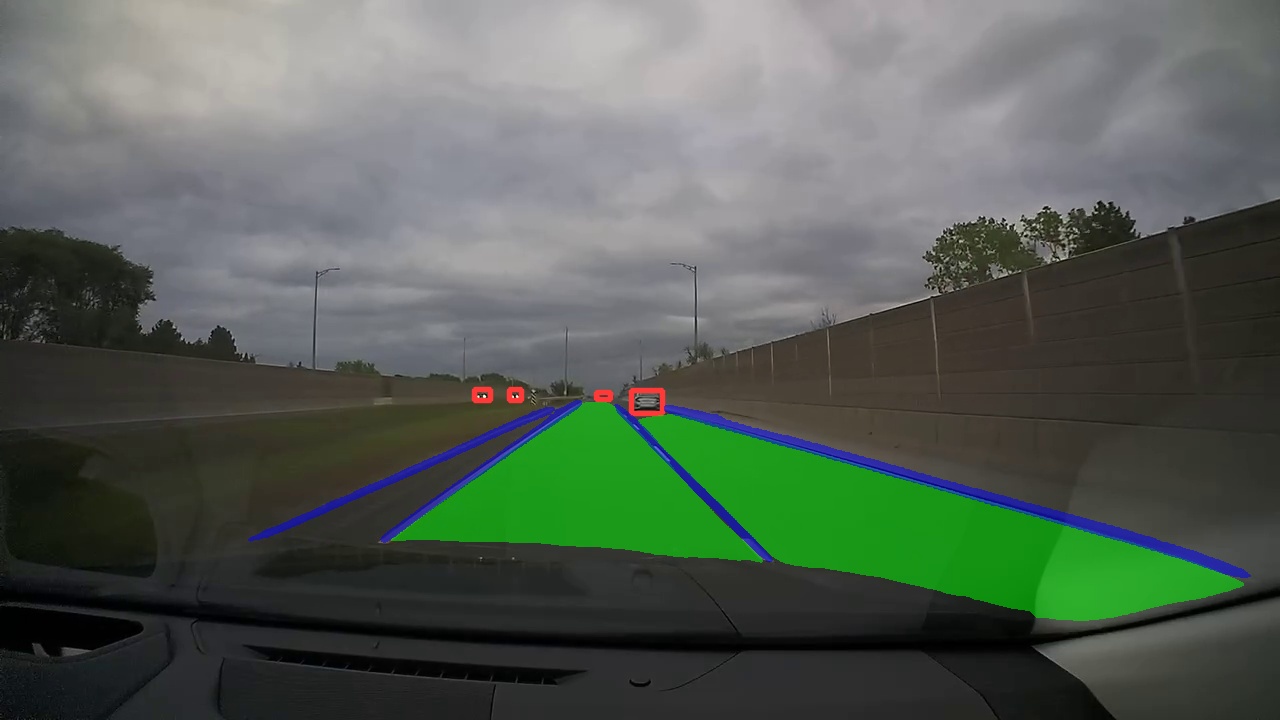}
    \end{subfigure}
    \caption{Real Road Results. Left image: nighttime. Middle image: daytime. Right image: highway.}
    \label{fig:real road}
\end{figure*}

\begin{table*}[!h]
\centering
\caption{The ablation study for the adaptive concatenation module}
\label{tab:training_methods_comparison}
\begin{tabularx}{\textwidth}{>{\centering\arraybackslash}X>{\centering\arraybackslash}X>{\centering\arraybackslash}X>{\centering\arraybackslash}X>{\centering\arraybackslash}X>{\centering\arraybackslash}X}
\toprule
Training method & Recall (\%) & mAP50 (\%) & mIoU (\%) & Accuracy (\%) & IoU (\%) \\
\midrule
YOLOM(n) & 85.2 & 77.7 & 90.6 & 80.8 & 26.7 \\
A-YOLOM(n) & 85.3 & 78 & 90.5 & 81.3 & 28.2 \\
YOLOM(s) & 86.9 & 81.1 & 90.9 & 83.9 & 28.2 \\
A-YOLOM(s) & 86.9 & 81.1 & 91 & 84.9 & 28.8 \\
\bottomrule
\end{tabularx}
\end{table*}

In this subsection, we present an ablation study to validate the effectiveness of our adaptive concatenate module. Additionally, we evaluate the impact of the segmentation neck and head on the overall model performance.

\subsubsection{Adaptive concatenation module}
\label{subsubsec: Adaptive concatenation module}
To assess the impact of our adaptive concatenation module, we compare performance with and without this module. Specifically, YOLOM(n) and YOLOM(s) as baselines, representing two distinct experiment groups with different backbones. We have carefully designed the segmentation neck structure for them to address the demands of particular segmentation tasks. However, both A-YOLOM(n) and A-YOLOM(s) include the adaptive concatenation module. Their segmentation neck structure is entirely identical and without well-design. The results are presented in Table~\ref{tab:training_methods_comparison}. When comparing A-YOLOM(n) with YOLOM(n), we find comparable performance in detection and drivable area segmentation. However, both lane line accuracy and IoU show significant improvements. Similarly, when comparing A-YOLOM(s) to YOLOM(s), we observe the same trend of improvement. These results indicate that our adaptive concatenation module could adaptively concatenate features without manual design and achieve a similar or better performance than well-designed segmentation heads, further enhancing the model’s generality.

\subsubsection{Multi-task model and segmentation structure}
\label{subsubsec: Multi-task model and segmentation structure}
To evaluate the influence of the multi-task approach for each individual task, as well as our proposed segmentation neck and head structures, we compare performance and head parameters in YOLOv8(segda), YOLOv8(segll), YOLOv8(multi), and YOLOM(n) within the domain of segmentation tasks. YOLOM(n) has a similar backbone network with YOLOv8, making this comparison fair and make sense. The results are presented in Table~\ref{tab:segmentation_comparison}. YOLOv8(segda) and YOLOv8(segll) implement the drivable area and lane line segmentation tasks separately. YOLOv8(multi) is an integrated multi-task learning model that combines YOLOv8(segda), YOLOv8(segll), and YOLOv8(det) neck and head structure into one shared backbone. We observe there is a significant performance improvement. This indicates that multi-task learning can reciprocally enhance the performance of individual tasks. 

\begin{table}[!h]
    \centering
    \caption{Results of different Multi-task model and segmentation structure}
    \label{tab:segmentation_comparison}
    \begin{tabularx}{\linewidth}{YYYYY}
    \toprule
    Model & Parameters & mIoU (\%) & \mbox{Accuracy (\%)} & IoU (\%) \\
    \midrule
    YOLOv8(segda) & 1004275 & 78.1 & - & - \\
    YOLOv8(segll) & 1004275 & - & 80.5 & 22.9 \\
    YOLOv8(multi) & 2008550 & 84.2 & 81.7 & 24.3 \\
    YOLOM(n) & 15880 & 90.6 & 80.8 & 26.7 \\
    \bottomrule
    \end{tabularx}
\end{table}

Compared to YOLOv8(multi), YOLOM(n) has our carefully designed neck structure tailored for segmentation tasks. Additionally, it boasts a significantly more lightweight head structure, having only 0.008 times the complexity. This significant improvement is due to our unique head design, which only relies on a series of convolutional layers, directly outputting a mask without the need for additional protos information. YOLOM(n) achieves impressive results in both mIoU and IoU for the drivable area and lane line tasks, respectively. Furthermore, we deliver competitive accuracy in the lane line task. These results evidence that our proposed neck and head innovations have achieved commendable results with minimal parameters and performance overhead.

\subsection{Real roads experiments}
\label{subsec: Real roads experiments}

This section primarily discusses the experiments conducted in real road datasets. Specifically, we capture several videos using a dash camera and then convert them frame by frame into images to predict using our model and YOLOP. Each converted image dataset consists of 1428 images with a resolution of 1280x720. These images encompass three scenarios: highway, nighttime, and daytime. Figure~\ref{fig:real road} displays the results from the real road dataset. However, our model consistently maintains relatively stable performance, particularly for the A-YOLOM(s). In all tasks, A-YOLOM(s) outperforms YOLOP. Meanwhile, A-YOLOM(n) is slightly inferior only in detection tasks compared to YOLOP. ADS must be capable of operating smoothly in unfamiliar scenarios. This is crucial. The results on real roads demonstrate that our model can meet the needs of autonomous vehicles on real roads.

\section{CONCLUSION}
\label{sec: CONCLUSION}

In this study, we primarily introduce an end-to-end lightweight multi-task model design for real-time autonomous driving applications. The advantages of the multi-task model lie in its capacity for each task to implicitly enhance the others, thereby further improving the performance of all tasks and enhancing the model's deployability on edge devices. To improve the model's generality, we integrate an adaptive concatenation module and propose unified loss functions for each type of task. This makes our model more flexible. When compared to other state-of-the-art real-time multi-task methods using the BDD100k dataset, our model not only demonstrates superior visualization results but also has a higher FPS. Moreover, real-road dataset evaluations prove our model's robustness in novel environments, positioning it ahead of the competition. In the future, we aspire to encapsulate more autonomous driving tasks within our model and optimize it for edge device deployments.

\bibliographystyle{IEEEtran}

\bibliography{ref}

\end{document}